\documentclass{article}

     \usepackage[preprint]{neurips_2022}

\usepackage[utf8]{inputenc} % allow utf-8 input
\usepackage[T1]{fontenc}    % use 8-bit T1 fonts
\usepackage{hyperref}       % hyperlinks
\usepackage{url}            % simple URL typesetting
\usepackage{booktabs}       % professional-quality tables
\usepackage{amsfonts}       % blackboard math symbols
\usepackage{nicefrac}       % compact symbols for 1/2, etc.
\usepackage{microtype}      % microtypography
\usepackage{xcolor}         % colors

\usepackage{amsmath}
\usepackage{amssymb}
\usepackage{subcaption}
\usepackage[inkscapepath=svgsubdir]{svg}
\usepackage{placeins}

\newtheorem{thm}{Theorem}
\newtheorem{prop}{Proposition}

\title{On the Interpretability of Regularisation for Neural Networks Through Model Gradient Similarity}

\author{%
  Vincent Szolnoky \\
  Department of Mathematical Sciences \\
  Chalmers University of Technology \\
  Göteborg, Chalmers Tvärgata 3, 41296, Sweden \\
  \texttt{szolnoky@chalmers.se} \\
  \And
  Viktor Andersson \\
  Department of Electrical Engineering \\
  Chalmers University of Technology \\
  Göteborg, Chalmersplatsen 4, 41296, Sweden \\
  \texttt{vikta@chalmers.se} \\
  \And
  Balázs Kulcsár \\
  Department of Electrical Engineering \\
  Chalmers University of Technology \\
  Göteborg, Chalmersplatsen 4, 41296, Sweden \\
  \texttt{kulscar@chalmers.se} \\
  \And
  Rebecka Jörnsten \\
  Department of Mathematical Sciences \\
  Chalmers University of Technology \\
  Göteborg, Chalmers Tvärgata 3, 41296, Sweden \\
  \texttt{jornsten@chalmers.se} \\
}

\begin{document}

\maketitle

%%%%%% Document begins

%KEYWORDS: neural network, regularisation, gradient descent, gradient similarity, noisy labels,

%TL;DR A new framework for monitoring and regularising neural network training,
%providing insights into the mechanism of regularisation for a wide range of methods

\section{Abstract}
Most complex machine learning and modelling techniques are prone to over-fitting and may subsequently generalise poorly to future data.
%Over-fitting often leads to a lack of generalisation and is a problem for many machine learning and complex modelling techniques.
Artificial neural networks are no different in this regard and, despite having a level of implicit regularisation when trained with gradient descent, often require the aid of explicit regularisers.
%To better understand regularisation of neural networks, and therein their generalisation properties, we introduce \emph{Model Gradient Similarity} (MGS).
We introduce a new framework, \emph{Model Gradient Similarity} (MGS),
% perhaps add a sentence here re thinking behind it
that (1) serves as a metric of regularisation, which can be used to monitor neural network training, (2) adds insight into how explicit regularisers, while derived from widely different principles, operate via the same mechanism underneath by increasing MGS, and (3) %forms
provides
the basis for a new
%class of -
regularisation scheme which exhibits excellent performance, especially in challenging settings such as high levels of label noise or limited sample sizes.
%To gain a better understanding of the regularisation of neural networks
%, and therein their generalisation properties,
%we introduce \emph{Model Gradient Similarity} (MGS).
%This new framework is based on the observation that the model gradients can be used to evaluate similarity between input data from the model's perspective. How similar a model deems input points during learning, and how that similarity evolves, ultimately has an effect on how the model generalises later on. When monitoring existing regularisation methods using metrics based on MGS, it is revealed that, despite them working differently on the surface, they operate via the same mechanism underneath by increasing MGS. %Motivated by this discovery, we develop a new type of regularisation scheme that directly optimises MGS during training. Thus, in addition to providing insight into the underlying mechanism of commonly used regulariser methods,
%MGS also works in a standalone manner. We demonstrate the top performance of MGS on several **** experimemts.
%when compared to other common explicit regularisers.
%MGS especially excels in the presence of high noise and/or limited training size.
%, MGS regularisation excels even further.
%This validates using MGS as a basis for evaluating %and designing a novel class of explicit regularisers.

\section{Introduction}
Since the inception of modern neural network architecture and training, many efforts have been made to understand their generalisation properties.
%and conditions.
When neural networks are trained with Gradient Descent (GD) this induces implicit regularisation in the network, causing it to attain surprising generalisation performance with no extra intervention. Despite this, neural networks will in many instances overfit in the presence of noise and have been shown to have the capacity to completely memorise data \citep{zhang16:_under}. Therefore, alongside the desire to understand how neural networks are implicitly regularised, explicit regularisation has been an ongoing pursuit to improve generalisation even further.

A broad range of explicit regularisers now exist that attack the problem from many different angles. %Following is a
Below, we provide a short overview of some of the most commonly used methods. \emph{Weight decay} \citep{krogh91:_simpl_weigh_decay_can_improv_gener} %is a common one that
places a L2-penalty on the parameters to encourage a minimum-norm solution.
%A number of normalisation schemes exist that
\emph{Normalisation schemes}, that normalise the data as it flows through the network, have been shown to act as explicit regularisers \citep{ioffe15:_batch_normal}. \emph{Double back-propagation} incorporates the gradients with respect to the loss itself as part of the loss, hence the name, as the gradient of the gradient will be evaluated in the complete back-propagation step \citep{drucker91:_doubl}. Although these methods were initially used as a part of a energy minimisation principle, they are now believed to help finding "flat minima", which have been shown to produce solutions that generalise better \citep{zhao22:_penal_gradien_norm_effic_improv}. \emph{Dropout}
%is a another common regularisation technique that
turns off connections between neurons during training and thus forces the network not to rely on any one given connection \citep{srivastava14:_dropout}. Another way to view the learning problem is to ignore the network itself and instead focus on the optimisation scheme. Here, a multitude of different learning rate and optimisers exist that offer explicit regularisation. Some standout examples include \emph{Cyclical Learning Rate} \citep{li20:_cyclic_learn_rate_method_deep_learn_train} and the \emph{Adam} optimisation algorithm \citep{kingma15:_adam}. More recently, especially %with the advent
in works related to the Neural Tangent Kernel (NTK) \citep{jacot18:_neural_tangen_kernel}, focus has been put onto functional regularisation. The core idea is to view the neural network architecture as being an opaque function approximator and instead regularise it as it is viewed from function space rather than the parameter space. This is usually achieved via a form of \emph{kernel ridge regression} or an approximation thereof \citep{bietti19:_kernel_persp_regul_deep_neural_networ,hoffman19:_robus_learn_jacob_regul}.

\textbf{Our contribution}. The fact that so many different explicit regularisers exist shows that their precise %function
mechanisms are not well understood. Ultimately, this leads back to the insufficient understanding of what causes neural networks to generalise in the first place. In this article, a new framework called \emph{Model Gradient Similarity} (MGS) is introduced. MGS builds on the idea that regularisation and, to a large
%larger
extent, generalisation, is tightly connected to how the model %itself
evolves during training. %One of the main driving forces determining how a model learns is %in fact
%its own gradients. It is therefore of interest to observe how the gradients evolve in relation to each other. %When MGS is used to analyse existing regularisers it is revealed that they, despite working very differently on the surface, appear to operate via the same underlying mechanism of increasing model gradient similarity.
We show that MGS can be used to analyse neural network training, monitoring how model gradients evolve together. It is revealed that explicit regularisers, despite their fundamentally different construction, operate via the same underlying mechanisms in terms of increasing model gradient similarity.
%When MGS is used to analyse existing regularisers it is revealed that they, despite their very different construct, appear to operate via the same underlying mechanism of increasing model gradient similarity.
%This gives strong evidence as to how neural networks can be holistically regularised.
%. %To that end,
This insight lays the foundation for a new class of regularisers aimed at directly boosting model gradient similarity.
%MGS is used to develop a new proof-of-concept regulariser that directly optimises a new penalty that aims to boost gradient similarity, showing that MGS works in a standalone manner.
We derive two simple MGS regularisation techniques from first principles and compare performance to a wide range of regularisation methods.
In a majority of the cases, the MGS regularisation achieves top performance and exhibits robustness qualities. This suggests that the MGS framework opens the door for the further development of both novel regularisers and the improvement of existing ones. %Finally, MGS also sheds some light on the implicit generalisation that comes from gradient descent.

\section{Model Gradient Similarity}
\label{sec:mgs}
Let \(x\in X\) denote an input observation in a data batch and \(y\in Y\) the target. We will train the model  \(f_{\theta}\), parameterised by \(\theta\), using loss function \(\mathcal{L}\) via gradient descent (GD). Thus,
the standard single-observation batch update rule using GD at time \(i\) is \(\theta_{i+1} = \theta_i - \eta \nabla_{\theta} \mathcal{L}\left( f(x), y \right)\), where \(\eta\) is the learning rate.
%For a given model \(f_{\theta}\), parameterised by \(\theta\), and loss function \(\mathcal{L}\),
%the standard update rule using GD is \(\theta_{i+1} = \theta_i - \eta \nabla_{\theta} \mathcal{L}\left( f(x), y \right)\) for some input data \(x\) and corresponding target data \(y\) at time step \(i\) with learning rate \(\eta\).
When training a neural network, \(f_{\theta}\) shall represent the raw output from the network before any additional normalisation is performed (e.g.\
%such as
softmax as in the case of classification).
%in the case of classification for example.

%Due to neural networks often exhibiting good generalisation performance when being trained with GD,
%The structure of
The loss gradients with respect to the parameters \(\nabla_{\theta} \mathcal{L}\) (\textbf{loss-parameter gradient})
%should studied as they
are thus the main contributing factor to the update of the model.  In the general case for common losses, such as mean-squared error or cross-entropy, we apply the chain rule to the loss-parameter gradient: \(\nabla_{\theta} \mathcal{L} = \nabla_{\theta} f_{\theta} \cdot \nabla_f \mathcal{L}\). We denote the two components as: the loss gradients with respect to the model output \(\nabla_f \mathcal{L}\) (\textbf{loss-model gradients}) and the model gradients with respect to the model parameters \(\nabla_{\theta} f_{\theta}\) (\textbf{model-parameter gradients}).

%To this end,
The idea of gradient similarity/coherence has  recently been suggested as an explanation for a neural network's ability to generalise \citep{faghri20:_study_gradien_varian_deep_learn,chatterjee22:_gener_myster_deep_learn}. It is here, however, mainly considered from the perspective of the loss-parameter gradient \(\nabla_{\theta} \mathcal{L}\) and not the model-parameter gradients \(\nabla_{\theta} f_{\theta}\). Contrary to the loss-parameter gradients, which diminish as any minima is approached, the model-parameter gradients display different behaviour and describe the evolution of the function realised by the model. To help understand why they can be useful in understanding neural network generalisation and regularisation thereof, we will borrow some inspiration about gradient similarity from \citet{charpiat19:_input_simil_neural_networ_persp} and use it to define the core ideas behind MGS.

\subsection{Defining MGS and its Relation to Generalisation}
When judging the similarity of two points \(x\) and \(x'\) from the model's point of view, one may be inclined to simply compare the output \(f_\theta(x)\) to the other \(f_\theta(x')\). Unfortunately, due to the non-linear nature of neural networks, the same output might be produced for a number of different reasons and it is therefore difficult to draw conclusions about similarity directly from the output.

An alternative notion of similarity between \(x\) and \(x'\) can be defined by how much changing the model output for \(x\) (by training on this sample) changes the output for \(x'\). If the points are dissimilar from the model's standpoint, then changing \(f_\theta(x)\) should have little influence over \(f_\theta(x')\) and vice-versa.
That is, after one step using GD for a single input sample \(x\) the parameters are changed by:
\begin{equation*}
  \delta \theta = -\eta \nabla_{\theta}f \nabla_f\mathcal{L}\left( f(x), y \right).
\end{equation*}

This induces the following change in the output of the model for that specific \(x\):
\begin{equation}
\label{eq:model-update-same}
\begin{aligned}
f_{\theta + \delta \theta}(x) &= f_\theta(x) + \nabla_\theta f(x) \cdot \delta\theta + O(||\delta\theta||^2) \\
&\approx f_\theta(x) -\eta  \left( \nabla_{\theta} f(x) \cdot \nabla_{\theta} f(x) \right) \nabla_f \mathcal{L}\left( f(x), y \right).
\end{aligned}
\end{equation}

On the other hand, this update will also yield a %similar
change in the model for another \(x'\) given by:
\begin{equation}
\label{eq:model-update-other}
\begin{aligned}
f_{\theta + \delta \theta}(x') &= f_\theta(x') + \nabla_\theta f(x') \cdot \delta\theta + O(||\delta\theta||^2) \\
&\approx f_\theta(x) -\eta \left( \nabla_{\theta} f(x') \cdot \nabla_{\theta} f(x) \right)\nabla_f \mathcal{L}\left( f(x), y  \right).
\end{aligned}
\end{equation}

\begin{figure}
  \centering
  \begin{subfigure}{0.49\textwidth}
    \includegraphics[width=\linewidth]{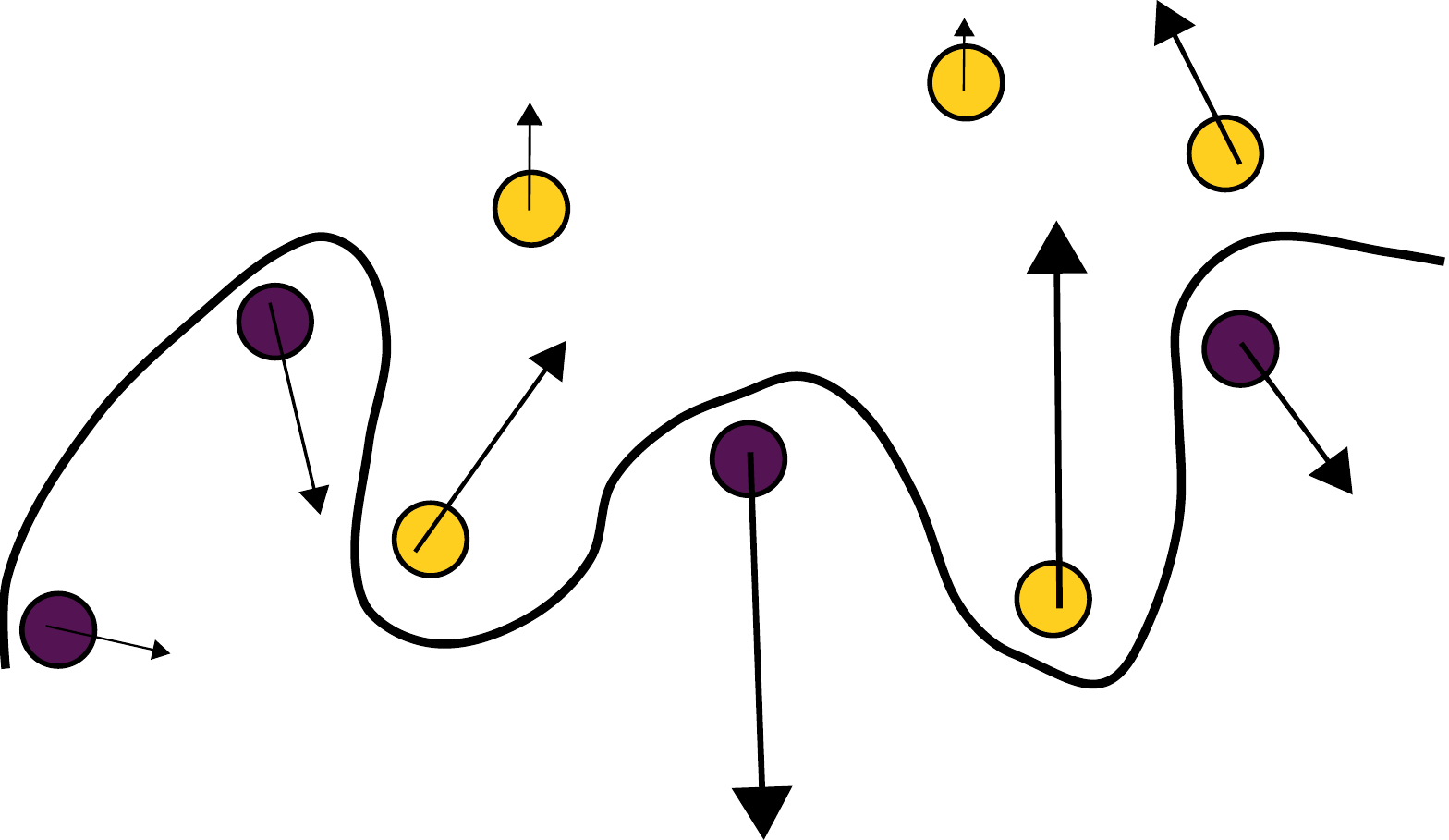}
    \caption*{Poor generalisation}
  \end{subfigure}
  \begin{subfigure}{0.49\textwidth}
      \includegraphics[width=\linewidth]{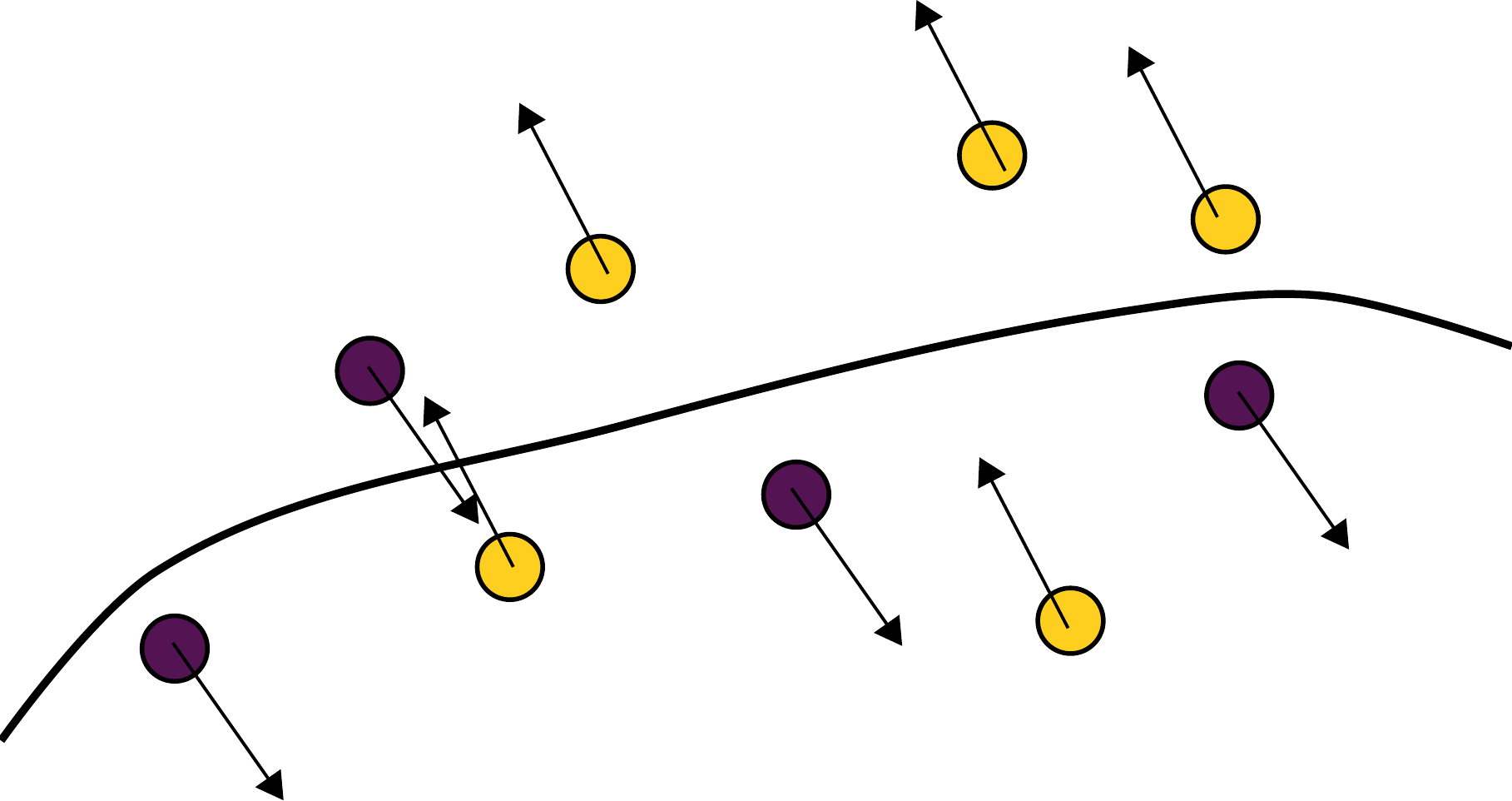}
      \caption*{Good generalisation}
  \end{subfigure}
  \caption{\label{fig:cartoon-example} {\bf An illustration on the connection between MGS and generalisation.} \  The arrows represent the {\bf model gradients}, \(\nabla f_{\theta} \), for each data point. {\bf Left:} A model that exhibits low similarity between its gradients. The model is free to adapt the gradients for each point and therefore learns them individually. The most likely outcome then is that it will overfit the data and generalise poorly. {\bf Right:} %Juxtaposed on the right is
  A model that is required to maintain a level of similarity between its gradients. To lower the overall loss, while maintaining model gradient similarity, the model is forced to learn from the data in groups defined by similar gradients.
  %where in each group the gradients do not differ that much.
  Thus, the model will have to learn separations in the data based on such groupings instead of each point individually.}
\end{figure}

Therefore, the kernel \(k_{\theta}(x, x') = \nabla_{\theta} f(x) \cdot \nabla_{\theta} f(x')\) (derived from the model-parameter gradients) is crucial to understanding how the model evolves after one step of GD. It determines how much of the loss-model gradient \(\nabla_f \mathcal{L}\left( f(x), y  \right)\) is used to update the model, up to a scaling by the learning rate. Another way to look at it is that \(k_{\theta}(x, x')\) describes the influence learning \(x\) has over \(x'\). If \(k_{\theta}(x, x')\) is large, which occurs when the model gradients are similar, then an update for \(f(x)\) will also move \(f(x')\) in the same direction. Likewise, if \(k(x, x')\) is small, meaning the model gradients are dissimilar, then the update for \(f(x)\) will have little affect on \(f(x')\). The cartoon example in figure \ref{fig:cartoon-example} illustrates how this property can be useful to understand how a model generalises better if it exhibits higher MGS.

\subsection{The contribution of MGS to Gradient Descent}
%Apart from determining to what extent the model gradients are aligned with each other, MGS also has another effect on gradient descent. Namely it causes the loss-model gradients for points that have high MGS with each other, to be averaged in the current step. Furthermore the average is weighted according to MGS which sets the factors.

If equations \ref{eq:model-update-same} and \ref{eq:model-update-other} above are combined, the complete update for a single step of gradient descent with training data \(X\), is:
\begin{equation*}
\begin{aligned}
\mathbf{f}_{\theta + \delta\theta}(X) &= \mathbf{f}_{\theta}(X) + \nabla_{\theta}\mathbf{f}(X) \cdot \delta\theta + O(||\delta\theta||^2) \\
&\approx \mathbf{f}_{\theta}(X) -\eta K_{\theta}(X) \cdot \nabla_f \mathcal{L}\left( \mathbf{f}(X)\right) \\
&= \begin{bmatrix}f_{\theta}(x_1) \\ \vdots \\ f_{\theta}(x_n) \end{bmatrix} -\eta \begin{bmatrix}k(x_1, x_1) & \cdots  & k(x_1, x_n) \\ \vdots & \ddots & \vdots \\ k(x_n, x_1) & \cdots & k(x_n, x_n) \end{bmatrix} \cdot \begin{bmatrix} \nabla_f \mathcal{L}(f(x_1), y_1) \\ \vdots \\ \nabla_f \mathcal{L}(f(x_n), y_n)\end{bmatrix}.
\end{aligned}
\end{equation*}

Here \(K_\theta(X) = k(X, X)\) is the kernel matrix for data \(X\).  The update for each model output \(f_\theta(x_k)\) can thus be seen as a weighted average
%between the kernel \(k(x_k, x_j)\) and '
of the loss-model gradients \(\nabla_f \mathcal{L}\left( f(x_j), y_j \right)\):
\begin{equation}
\label{eq:model-update-average}
f_{\theta + \delta\theta}(x_k) = f_{\theta}(x_k) - \eta \sum_{j=1}^n k(x_k, x_j) \cdot \nabla_f \mathcal{L}(f(x_j), y_j),\ x_k \in X.
\end{equation}

From equation \ref{eq:model-update-average}, it follows that the kernel ultimately controls the weighting of the loss-model gradients: the greater the similarity exhibited between the model gradients, the greater the averaging effect will be. This partially explains how GD induces implicit regularisation as, unless \(K_\theta(X)\) is dominated by its diagonal, a model update for observation $x_k$ will also utilise the loss gradients of other observations.
%there will always exist some amount of averaging.

Furthermore, it has been observed from the Neural Tangent Theory perspective that \(K_\theta(X)\) will often align with the ideal kernel \(YY^T\) in classification problems, which perfectly discriminates targets. This phenomenon has already been investigated as one of the reasons behind implicit generalisation for neural networks when trained with SGD \citep{kopitkov20:_neural_spect_align,baratin21:_implic_regul_neural_featur_align}. Here, we also
%find a simple reason as
see that the alignment will naturally engage the averaging effect, grouping gradients according to their target classes. It is still unclear what the cause of the alignment is. However, MGS provides an explanation as to why such an alignment is useful. More detail on the connection to Neural Tangent Theory is presented in the supplementary material.

To summarise, apart from capturing to what extent the model gradients are aligned with each other, %MGS also has another effect on gradient descent. Namely it causes
MGS also determines how loss-model gradients for observations that have high MGS with each other are averaged in the GD step.
%Furthermore the average is weighted according to MGS which sets the factors

% \subsection{The Effect of MGS on Generalisation}
% From a generalisation/regularisation perspective this notion of model gradient similarity should be useful as illustrated in the cartoon example in figure \ref{fig:shared-gradients}. The less influence changing the function output with regards to a selected input has on others the more likely it is that it's being overfit/memorised by the network. That is to say the network is free to learn that particular input without affecting its learning of any of the others. On the other hand, if the opposite is true and changing the function output for a single input drastically changes the outputs of others, then it is likely the network thinks of these inputs as a group that is similar. In the extreme case where changing one output affects a majority of others might be as a result of underfitting.

\subsection{Metrics to Quantify MGS}
\label{sec:useful-metrics}
%Given the previously defined
Since the kernel \(K_{\theta}(X)\) captures coordinated learning between similar observations (equation \ref{eq:model-update-average}), it is desirable to identify metrics which can summarise the overall model gradient similarity in a batch of data.

To find relevant metrics of \(K_\theta\)
we note that
%that could be used it's first helpful to identify that that
the spectrum of \(K_{\theta}(X) = (\nabla_\theta f(X))^T \nabla_\theta f(X)\) is the same as that of \(\tilde{C} = \nabla_\theta f(X) (\nabla_\theta f(X))^T\);
%Additionally \(\widehat{C}\) is
the non-centered version of the sample covariance matrix \(C = (\nabla_\theta f(X) - \overline{\nabla_\theta f(X)}) (\nabla_\theta f(X) - \overline{\nabla_\theta f(X)})^T\). The spectrum of \(C\) and \(\tilde{C}\) are interlaced as their corresponding eigenvalues \(\lambda_1 \leq \tilde{\lambda}_1 \leq \ldots \leq \lambda_n \leq \tilde{\lambda}_n\).

We also note that the trace and determinant of \(C\) are commonly used measures of the overall variance and covariance for a given sample of data. Here, the model-parameter gradients \(\nabla f_{\theta}\) constitute the data. Moreover, the trace and determinant of \(\hat{C}\) give a rough approximation of the same metrics of \(C\). Proofs of these two properties are given in the supplementary material.

%The trace and determinant of \(C\) are commonly used measures of the overall variance and covariance for a given sample of data. Here, the model-parameter gradients constitutes the data.

Thus, the summary statistics
% interpretation of
%\begin{equation*}
%\operatorname{tr}K_{\theta} \text{\ \ \ and \ }
%\operatorname{det}K_{\theta}
%\end{equation*}
\(\operatorname{tr} K_{\theta}\) and \(\operatorname{det} K_{\theta}\)
can be seen as approximations of the model gradient variance and covariance.
%which
%in turn gives an idea on the overall similarity between gradients.
\emph{Smaller} values for the trace and determinant of the kernel, \(K_{\theta}\), reflect a higher degree of model gradient similarity and thus a larger averaging effect in equation \ref{eq:model-update-average}.

These metrics also serve as proxies for measuring the magnitude of the diagonal elements and relative magnitude of the diagonal-to-off-diagonal elements of \(K_{\theta}\).
From this perspective, smaller values for the trace and determinant of the \(K_{\theta}\) reflect how well the kernel can be summarised by a low-rank representation. Low-rank \(K_{\theta}\) can be viewed as indicative of coordinated learning in that it restricts the number of independent directions in which the functions can evolve in equation \ref{eq:model-update-average}.

%These metrics also appear in a wide range of fields. For low-rank matrix %approximations (e.g.\ \cite{candes2009exact}), the trace serves as a proxy for low-rank which can here be interpreted as coordinated learning, limiting the independent directions in which the model can evolve. In MORE HERE e.g. stability....

%constitutes measures of overall similarity between gradients. The exact values of these metrics are not important. Rather, we wish to monitor how they evolve during training and so an approximation suffices.

From a practical standpoint, the exact values of these metrics are not important. Rather, we wish to monitor how they evolve during training. Utilising the kernel also constitutes a pragmatic solution since calculating \(C\), of size \( p \times p \),  %any meagerly large neural network
is near infeasible for larger networks, whereas
%as it requires \(C\) which is \(p^2\) in size versus
\(K_{\theta}\) is only of size \(n \times n\), where \(p\) and \(n\) are the number of parameters and data batch respectively, and generally \(n \ll p\).

\section{Tracking MGS Metrics During Training}
\label{sec:observing-mgs}
%Using the previously defined metrics,
Let us now investigate how the metrics \(\operatorname{tr} K_{\theta}\) and \(\operatorname{det} K_{\theta}\)
evolve during training. Using these metrics, a wide range of regularisers, representing the most commonly used ones, are compared.
%their evolution is tracked in some test examples, to see if they provide any information. A number of different regularisation methods are tested, representing the most commonly used ones.
Additionally, two new regularisation methods based on MGS are plotted alongside as a reference. These are introduced in the next section.

\begin{figure}
  \centering
  \includegraphics[width=\linewidth]{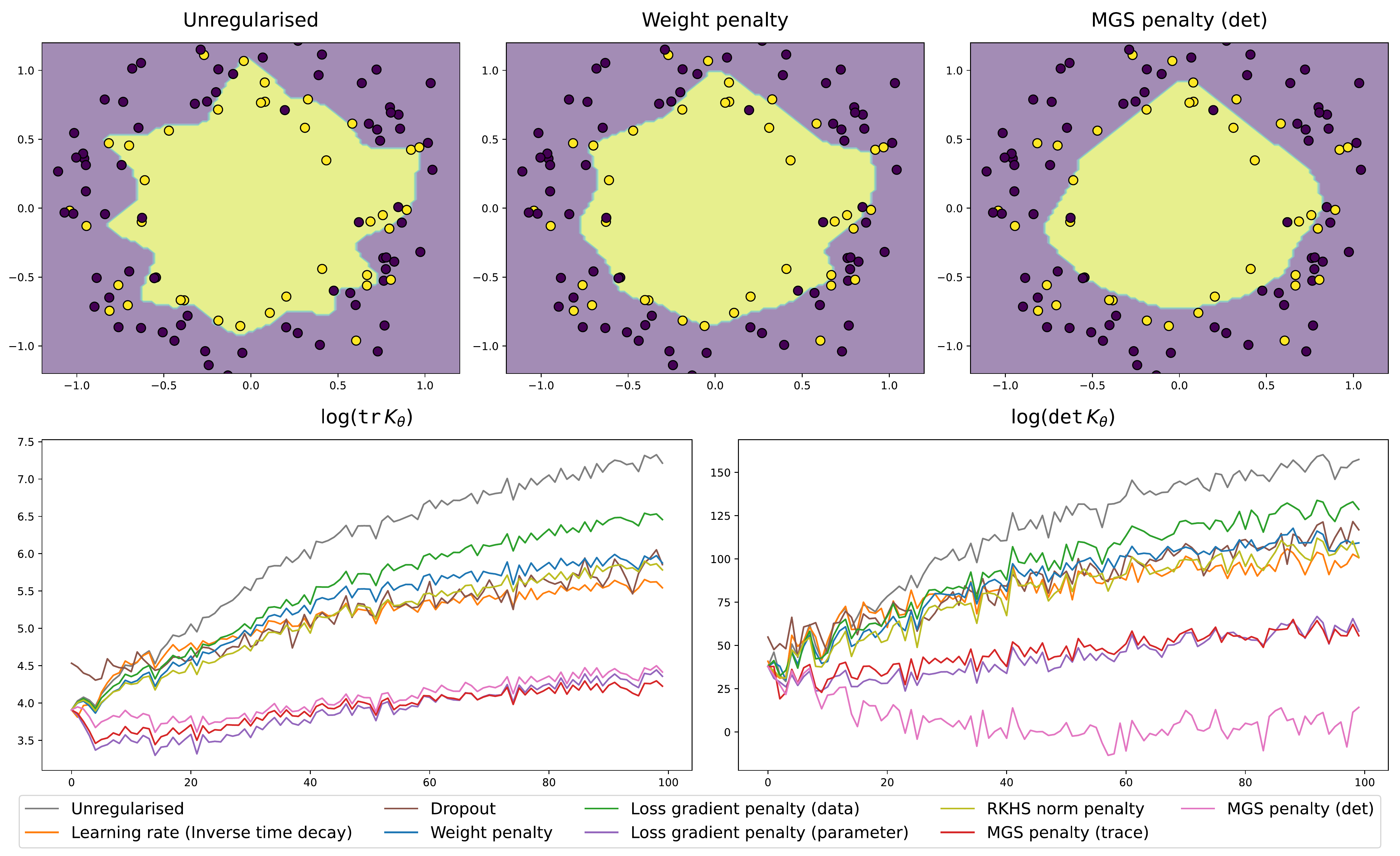}
  \caption{\label{fig:two-circles-example} {\bf FCN on two-circle classification.} \ {\bf Top: } Decision boundaries are shown to visualise how well the network has generalised for each method. {\bf Bottom:} The MGS metrics for each regulariser are tracked during the training period.
  The connection between network generalisation and the MGS metrics is clear.
  %There is a obvious connection between how well the network has generalised and the evolution of the MGS metrics.
  For an unregularised network which overfits the data, MGS metrics grow rapidly. Once any explicit regularisation is used, MGS metric growth is slowed. Coupled with this, the decision boundaries approach the true model for methods that constrain the MGS metrics better. (Complete set of decision boundaries are provided in the supplementary material.)}
\end{figure}

\begin{figure}
  \centering
  \includegraphics[width=\linewidth]{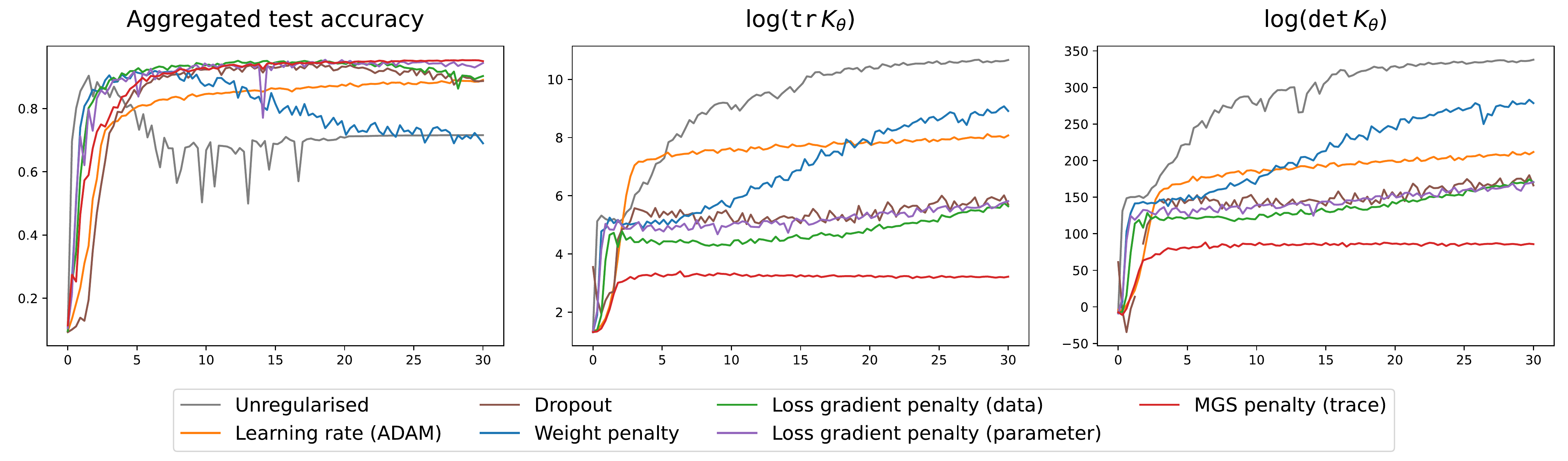}
  \caption{\label{fig:mnist-example}
  {\bf AlexNet on Corrupted MNIST.} {\bf Left:} Test accuracy. {\bf Middle and Right:} MGS metrics. Test accuracy is clearly coupled to MGS evolution.
   All methods that yield high final test accuracy exhibit small and slow-increasing MGS metrics, indicating a higher degree of model gradient similarity.}
  %\textbf{Example showing training on MNIST.} Similar to figure \ref{fig:two-circles-example} but now for MNIST, the test accuracy is plotted to guage how well the network generalises, alongside the MGS metrics. \textbf{Left:}
\end{figure}

The results are shown in figures \ref{fig:two-circles-example} and \ref{fig:mnist-example}.
%The former is
Figure \ref{fig:two-circles-example} presents results from a simple FCN architecture trained on a toy problem generated from two concentric cirles perturbed with noise. Figure \ref{fig:mnist-example} stem from training a large convolution network (AlexNet) on a corrupted version of MNIST with label noise. In each example, despite being different problems and different architectures, all regularisation methods exhibit the same behaviour: when regularisation is applied the rate of MGS metric growth is decreased, indicating higher model gradient similarity. Furthermore, the test accuracy or generalisation performance (figure \ref{fig:mnist-example}) is strongly correlated with the evolution of the MGS metrics.
%there even looks to be a correlation between how well the network generalises and the evolution of the MGS metrics.
The slower the rate of MGS growth, the better the final generalisation, and when the growth plateaus so does the test accuracy.
%to some degree.

%Here are also some interesting observations specifically from the more complex
We list some interesting observations from the
MNIST example in figure \ref{fig:mnist-example}:
\begin{itemize}
\item{The unregularised network (gray) goes through a rapid boost in accuracy and then quickly overfits. When test accuracy peaks, there is a small plateau in the MGS metrics. However, once  it starts to decline in accuracy, the MGS metrics again grow, indicating model gradient similarity is decreased.}
\item{Weight penalty (blue) is able to regularise the network initially, but also eventually overfits.
%at a slow rate.
This is reflected in the MGS metrics as
%once again
they plateau when test accuracy peaks and then start to increase at a seemingly proportional rate to the decline in performance.}
\item{All methods that achieve high test accuracy
%maintain
control the growth of the MGS metrics.
%a slow rate.
Still, many of them begin to overfit towards the end of training
%and this is seen in the
which coincides with the MGS metrics also gradually increasing.}
\item{Only the MGS penalty (to be introduced in the next section) is able to maintain a stable test accuracy
%by not letting the MGS metrics increase
}
\item{Dropout (brown) has gaps in its determinant metric. This means that it is causing \(K_{\theta}\) to be singular which could have ramifications on the stability of the training.}
\end{itemize}
More extensive test bench experiments are provided in section  \ref{section:experiments}.

\section{MGS regularisation}
%Optimising MGS Directly}
%With the current observations from
In the previous section, we saw a clear connection between boosting MGS (reflected by lower MGS metrics) and the ability of the network to generalise. The effect on MGS by the explicit regularisers was common to all the methods despite their widely differing design. Thus, we may think of
%It seems like
each
regularisation method as acting as a proxy for enforcing model gradient similarity. This motivates the construction of a new regularisation scheme that directly optimises MGS.

%To do this we shall
We thus modify the original loss to include a new penalty term \(g(K_\theta(X)))\) which acts on the MGS kernel:
\begin{equation*}
\widehat{\mathcal{L}}(f(X), Y) = \mathcal{L}(f(X), Y) + g(K_{\theta}(X))
\end{equation*}

We already have two prime candidates for \(g\), namely the previously investigated MGS metrics:
\begin{align*}
g_{\operatorname{tr}}(K_\theta(X)) &= \alpha \operatorname{tr}K_{\theta}(X) = \alpha \sum_i \lambda_i \\
g_{\operatorname{det}}(K_\theta(X)) &= \alpha \operatorname{det}K_{\theta}(X) = \alpha \log \left(\prod_i \lambda_i\right),
\end{align*}
where \(\alpha\) is a penalty factor and \(\lambda_i\) are the eigenvalues from the spectral decomposition of \(K_\theta\).
As previously mentioned in section \ref{sec:useful-metrics}, smaller values for these penalty metrics reflect higher model gradient similarity. Adding an explicit penalty on these metrics thus forces the network to learn data in groups and preventing it from learning points individually (over-fitting or memorization).

The evolution of the two were shown in figures \ref{fig:two-circles-example} and \ref{fig:mnist-example}. The metrics seem to behave similarly. However, \(g_{\operatorname{tr}}(K_{\theta})\) is more efficient to compute as it does not require either the complete spectral decomposition nor the full gradient similarity kernel to be computed. However, since both metrics are gradient based, they are, of course, computationally more burdensome than commonly used regularisers such as weight decay. For calculations of the penalties we make use of \emph{Neural Tangents} \citep{novak19:_neural_tangen} and \emph{Fast Finite Width NTK} \citep{novak22:_fast_finit_width_neural_tangen_kernel}  libraries which are built on \emph{JAX} [\cite{bradbury18:_jax}].

We note that the penalties are equivalent to penalising the arithmetic and geometric mean of the eigenvalues of the kernel, respectively. The former will mainly be affected by large leading eigenvalues. The latter essentially penalises the eigenvalues of the kernel on a log-scale and may thus take more of the overall structure of the kernel into account. Supplemental figure A.1 depicts results on the toy two-circle data from using both penalties. The differences are subtle. For pragmatic considerations, on real data we therefore utilise only the trace penalty for explicit regularisation, but both metrics can be used for monitoring neural network training.

%The test accuracy performances of the two penalties on simple data examples are shown in the appendix.

%\ref{fig:two-circles-example} and \ref{fig:mnist-example}. Both metrics seem to behave similarly and offer similar performance, however \(g_{\operatorname{tr}}K_{\theta}\) is more efficient to compute as it does not require either the complete spectral decomposition nor the full gradient similarity kernel to be computed. However, these penalties are equivalent to penalizing the arithmetic and geometric mean of the eigenvalues of the kernel, respectively. The former will mainly be affected by large leading eigenvalues while the latter
%will be more appropiate in these circumstances.

%As previously mentioned in section \ref{sec:useful-metrics}, these act equivalent to the arithmetic and geometric mean of the eigenvalues. The former will be affected by large leading eigenvalues while the latter will be more appropiate in these circumstances.

%By minimising these two penalties, model gradient similarity is increased, forcing the network to learn data in groups and preventing it from learning points individually (over-fitting or memorization).

%The test accuracy performances of the two penalties on simple data examples are shown in figures \ref{fig:two-circles-example} and \ref{fig:mnist-example}. Both metrics seem to behave similarly and offer similar performance, however \(g_{\operatorname{tr}}K_{\theta}\) is more efficient to compute as it does not require either the complete spectral decomposition nor the full gradient similarity kernel to be computed.

Details on how $K_{\theta}$ is calculated with regards to mini-batches, vector-outputs and large data sets are provided in the supplementary material.

\section{Experiments}
\label{section:experiments}

We compare the performance of optimising MGS directly versus the most common, explicit regularisers. Although there exists a plethora of explicit regularisers and modifications thereof, the ones chosen are a representative of the main types of regularisers with the most widespread use in practice. As the aim is to compare performance between regularisers, a systematic investigation is done where the regularisers are rigorously tested against common overfitting scenarios caused by target noise and training size.
%As many scenarios have to be run, the datasets are chosen that do not require overly complex networks to achieve reasonable performance on.
Two main tests are done, one on a classification problem and the other on a regression problem. For each test two different types of architectures are chosen. This is to test how universal each method is in handling vastly different problems and networks. Finally, a test of robustness to variability found in common training setups is performed. Complete details on the experiments, complete results and code to run them can be found in the supplementary material.

\subsection{Classification: corrupted MNIST}
We generate a corrupted version of the popular MNIST dataset by applying a motion blur to each image. A testbench of challenging scenarios are created by varying the amount of label noise and training size. Each regularisation method is tuned in an identical setting and then retains the chosen parameters for the entire testbench. This ensures an even playing field and that each method is given the same starting point. For each testing scenario, multiple runs are performed using a new network initialisation and resampling of the training data. %with the final test accuracy results being aggregated.

Table \ref{tbl:perf-mnist-lenet} shows the results for 4 scenarios from the 143 scenario large testbench (see supplemental). In this scenario 3000 samples are used for training at 4 noise levels. It's immediately clear that MGS achieves the best performance overall in many aspects. Not only does it reach higher accuracy levels than the other methods, it's also robust in its performance. This can be seen both by the standard deviation of its final test accuracy, but also when comparing the max test accuracy to its final one. For MGS max and final test accuracy essentially coincide, indicating that MGS regularised networks do not overfit. The same conclusions hold true for a fully connected network (see supplemental) where performance is overall worse for all methods but MGS outperforms the others.

\begin{table}
  \caption{\textbf{Test accuracy for corrupted MNIST dataset using a LeNet-5 architecture.} Final test accuracy and one standard deviation is shown with the maximum test accuracy in parenthesis underneath.
  Noise column represents percentage of training labels that have been randomly flipped. Note the ability of MGS to handle large amounts of noise. Comparing max accuracy (in parenthesis) during training to final accuracy, MGS is also the most consistent and is not susceptible to overfitting.}
  \label{tbl:perf-mnist-lenet}
  \centering
  \begin{tabular}{cccccc}
    \toprule
    % \multicolumn{5}{c}{Gradient penalty (parameter)}                   \\
    % \cmidrule(r){1-2}
    % Name     & Description     & Size ($\mu$m) \\
    Noise & Unregularised & Dropout & Weight & Loss grad. & MGS \\
    \midrule
    0\% & \parbox{5em}{\centering 89.6 \textpm 1.3 \\ (90.2)} & \parbox{5em}{\centering \textbf{95.8} \textpm 0.5 \\ (95.9)} & \parbox{5em}{\centering 83.1 \textpm 9.6 \\ (87.1)} & \parbox{5em}{\centering 95.1 \textpm 0.5 \\ (95.1)} & \parbox{5em}{\centering 95.2 \textpm 0.3 \\ (95.2)} \\
    \addlinespace
    30\% & \parbox{5em}{\centering 62.0 \textpm 2.5 \\ (85.2)} & \parbox{5em}{\centering 80.9 \textpm 2.9 \\ (92.0)} & \parbox{5em}{\centering 72.2 \textpm 4.2 \\ (79.2)} & \parbox{5em}{\centering 88.7 \textpm 2.0 \\ (92.0)} & \parbox{5em}{\centering \textbf{93.1} \textpm 0.7 \\ (93.4)}\\
    \addlinespace
    60\% & \parbox{5em}{\centering 37.3 \textpm 2.9 \\ (73.7)} & \parbox{5em}{\centering 59.2 \textpm 4.1 \\ (83.7)} & \parbox{5em}{\centering 10.0 \textpm 0.4 \\ (62.3)} & \parbox{5em}{\centering 80.7 \textpm 5.9 \\ (81.4)} & \parbox{5em}{\centering \textbf{88.4} \textpm 1.4 \\ (89.1)} \\
    \addlinespace
    80\% & \parbox{5em}{\centering 23.9 \textpm 2.4 \\ (62.4)} & \parbox{5em}{\centering 39.5 \textpm 4.7 \\ (56.6)} & \parbox{5em}{\centering 10.1 \textpm 0.5 \\ (23.5)} & \parbox{5em}{\centering 15.1 \textpm 5.4 \\ (17.2)} & \parbox{5em}{\centering \textbf{74.5} \textpm 4.0 \\ (75.1)} \\
  \bottomrule
  \end{tabular}
\end{table}

\subsection{Regression: Facial Keypoints}
A similarly corrupted version of the Facial Keypoints data set was used as a regression benchmark. %In this instance an entire grid was not run, and
Based on images of human faces, the problem is to predict the coordinates of 15 key facial features. Each image is first corrupted using a motion blur. Noise is then introduced by adding normally distributed values, using different scale parameters, to the target variables. Here, we fixed the number of training samples while the amount of target noise was varied. Each method is tuned using the same scenario first. Similar conclusions can be drawn as from the classification results. The results are visible in table \ref{tbl:perf-fk-lenet} where the noise column corresponds to the scale parameter of the normally distributed noise. Interestingly, only Dropout and MGS were actually able to converge to a satisfactory performance level. However, when results are compared for a FCN architecture  (table \ref{tbl:perf-fk-fcn}, supplemental), Dropout falls into the same category as the other methods while MGS performance remains high. This shows a weakness in a method such as Dropout: that it is inevitably architecture dependent.
%On the other hand MGS does not suffer from this problem.

\begin{table}
  \caption{\textbf{Test loss for the corrupted Facial Keypoints dataset using a LeNet-5 architecture.} Final test loss and one standard deviation is shown with the minimum test loss in parenthesis underneath. Only Dropout and MGS achieve an acceptable level of final test loss. The other methods also provide a marginal increase in performance compared to the unregularised network. In the supplemental we compare results on a FCN architecture for which only MGS is able to attain good performance.}
  \label{tbl:perf-fk-lenet}
  \centering
  \begin{tabular}{cccccc}
    \toprule
    Noise & Unregularised & Dropout & Weight & Loss grad. & MGS\\
    \midrule
    0 & \parbox{5.2em}{\centering 68.2 \textpm 36.7 \\ (54.2)} & \parbox{5.2em}{\centering 14.9 \textpm 2.2 \\ (13.9)} & \parbox{5.2em}{\centering 113.2 \textpm 53.2 \\ (90.3)} & \parbox{5.2em}{\centering 212.4 \textpm 174.1 \\ (104.0)} & \parbox{5.2em}{\centering \textbf{14.1} \textpm 1.0 \\ (13.8)}\\
    \addlinespace
    10 & \parbox{5.2em}{\centering 54.8 \textpm 24.0 \\ (45.2)} & \parbox{5.2em}{\centering 15.4 \textpm 2.2 \\ (14.4)} & \parbox{5.2em}{\centering 92.6 \textpm 46.5 \\ (69.0)} & \parbox{5.2em}{\centering 235.2 \textpm 188.6 \\ (140.5)} & \parbox{5.2em}{\centering \textbf{14.3} \textpm 1.1 \\ (13.6)}\\
    \addlinespace
    20 & \parbox{5.2em}{\centering 102.0 \textpm 116.2 \\ (48.2)} & \parbox{5.2em}{\centering 16.6 \textpm 2.2 \\ (15.3)} & \parbox{5.2em}{\centering 83.8 \textpm 52.7 \\ (49.3)} & \parbox{5.2em}{\centering 266.1 \textpm 233.1 \\ (115.9)} & \parbox{5.2em}{\centering \textbf{15.3} \textpm 1.3 \\ (14.7)}\\
    \addlinespace
    30 & \parbox{5.2em}{\centering 94.8 \textpm 131.5 \\ (39.0)} & \parbox{5.2em}{\centering 19.7 \textpm 2.8 \\ (17.7)} & \parbox{5.2em}{\centering 70.6 \textpm 58.0 \\ (41.9)} & \parbox{5.2em}{\centering 229.4 \textpm 225.7 \\ (151.5)} & \parbox{5.2em}{\centering \textbf{17.8} \textpm 2.0 \\ (16.9)}\\
    \bottomrule
  \end{tabular}
\end{table}

\subsection{Training parameter robustness}
\begin{figure}
  \centering
  \includegraphics[width=\linewidth]{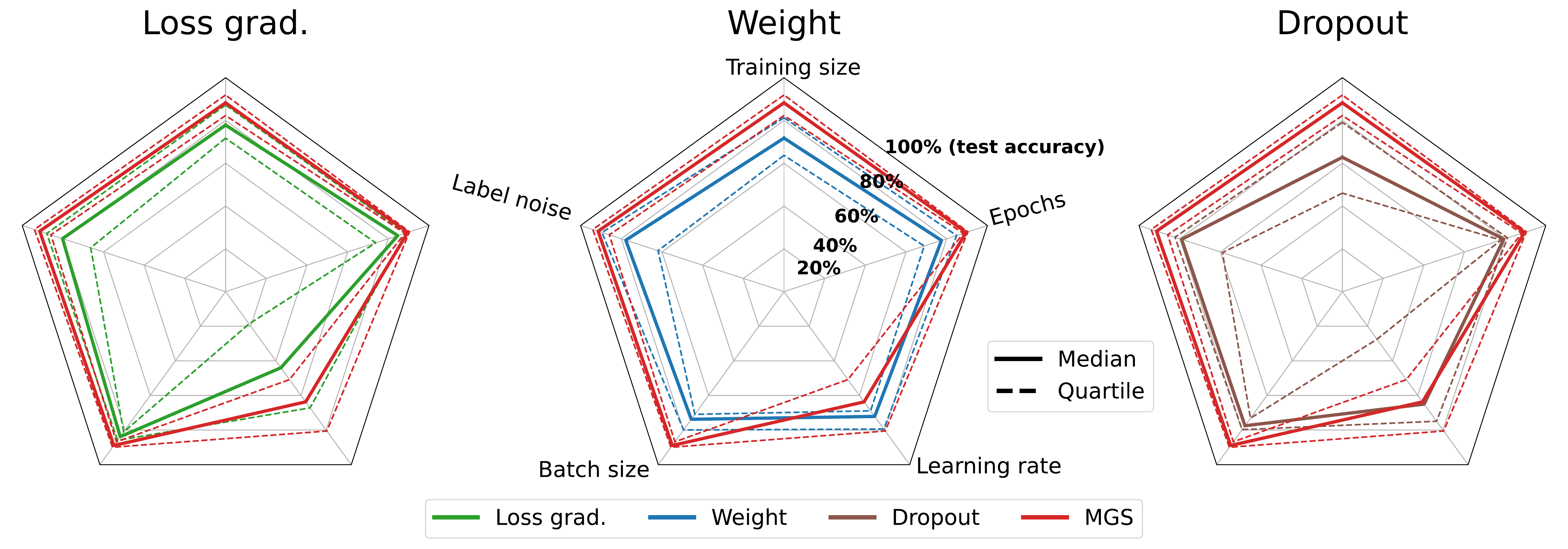}
  \caption{\textbf{Test accuracy quantiles after varying different parameters controlling training.} MGS is plotted in red against the other methods. MGS
  %stills
  %has better performance than the median for all the other methods.
  outperforms the other regularisers, and is also the most robust with regards to changes in the training setup, in all but one case.
  %With varying learning rate weight decay is somewhat better
 % MGS is also the most robust with regards to changes in the training setup and doesn't vary in its performance.
 The only area in which it shows a degradation in performance is when the learning rate is changed. However this seems to affect all methods, but weight penalty to a lesser degree.}
   \label{fig:radarplots}
\end{figure}

Finally, we test the robustness of each regularisation method by changing: amount of training size, label noise, batch size, learning rate, and epochs. A middle-ground scenario was chosen from this MNIST testbench as a starting point to tune each method. Then, the five training variables were changed individually to both higher and lower values, tracking the performance of each method after training. Consistent with the current findings, MGS outperforms each method substantially. In figure \ref{fig:radarplots} we use radar charts to summarise the test bench results.

% NOTE: Jag har inte lagt till dessa i supplemental.
%In the supplemental, the results are presented in full.

%All of these parameters, apart from label noise, are for the most part arbitrarily set before training. Even label noise can be seen as a training variable, as how the data is retrieved, preprocessed, sampled, etc, can in most cases be controlled.
We note that MGS is the least affected by a change in all but one of the test parameters or conditions, and exhibits superior performance compared to the other regularisers.
%If changing such simple parameters, yields differing performance in the methods ability to regularise, it brings into question how effective they actually are. In this regard, MGS is the most generic, and
Performance is only affected by learning rate to some extent. This is not unexpected as the same can be seen for the other methods apart from weight penalty. Also, as is evident from the GD update step (equation \ref{eq:model-update-same}), the one directly contributing training parameter in MGS is the learning rate.

% Additionally two common but different network architectures are chosen for both problems: a moderately deep fully connected network (FCN that is 6 hidden layers deep and 300 neurons wide) and a small convolutional network (LeNet-5). Each regularisation method is first tuned via means of a grid search in a setting with high label noise and small training set so as to ensure that they are properly calibrated. The final parameters are then used for the entire test-bench.

% For a regularisation method to be ideal it should not enforce any regularisation when the problem does not require any, for example in the absence of noisy data. Therefore it is preferred to tune for the harder scenarios rather than the easy ones. Each scenario is tested using multiple runs by shuffling and sampling data with the results being aggregated. Below we present a small selection of the results with a more comprehensive suite is provided in supplementary material with further explanations.

\section{Conclusion}
%XXX FW: computation, clustering effect, MGS is the most generic/holistic method

% GLÖMN INTE COMPUTATIONAL INTENSITY

We introduced the concept of \emph{Model Gradient Similarity} (MGS) and discussed its connections to regularisation for models trained with gradient descent and the generalisation properties of neural networks. We proposed metrics that can be used to summarise MGS for a model and track the training of different neural network architectures for various learning problems.

It was shown that a wide range of explicit regularisers all appeared to attempt to enforce higher model gradient similarity, i.e.\ lower MGS metrics. Moreover, higher test accuracy performance was shown to be reflected in lower MGS metric values.

%This provided insight into how regularisers function generically when training neural networks, but also how neural networks are implicitly regularised when trained with GD.

%When regularisation was used, regardless of the actual method, it was observed that each scheme attempted to enforce higher MGS. This could prove key in not only understanding how regularisers function generically when training neural networks, but also how neural networks are implicitly regularised when trained with GD.

Based on these findings, a new type of regulariser, geared toward direct control of MGS,  was designed and found to achieve top performance in several rigorous test bench experiments. Its overall robustness to label noise and training parameter settings was also an indication that directly optimising MGS comes closer to a more holistic approach to regularisation.

Taken together, these results provided insight into the underlying mechanisms of neural network regularisation. Due to the higher computational costs for gradient based regularisers, such as the MGS metric penalities introduced here or loss-gradient penalties, their use for direct optimisation is not always efficient. To scale for use in larger networks and in more complex settings, additional work is needed to obtain more efficient ways to compute or approximate the MGS kernel or metrics thereof. Future work could thus focus both on how MGS can be used to design new regularisers as well as to improve upon existing ones. The MGS metrics can be useful as KPI's for measuring a network's current capacity for under/over-fitting. Finally, the grouping effect of regularised neural network training, where model gradient similarity encourages coordinated learning across observations, suggests that MGS regularisation can be explored for joint prediction modeling and clustering.

%### NEEDS SOME WORK ###

%The main conclusion being, that while these methods based directly on MGS work, they are merely a first iteration. Instead future work should be focused on how MGS can be used to both design new regularisers that boost MGS directly but also better understand and improve existing ones. For instance, the metrics introduced can be useful as KPI:s for measuring a network's current capacity for under/overfitting.

\begin{ack}
This research was supported by funding from Centiro Solutions, the Swedish Research Council (VR), the Swedish Foundation for Strategic Research, and the Wallenberg AI, Autonomous Systems and Software Program (WASP).
\end{ack}

%%%%%% Document ends

\newpage

\bibliography{first_paper.bib}

\newpage

\appendix

% NOTE: Recent figure counter
\renewcommand{\thetable}{A.\arabic{figure}}
\renewcommand{\thefigure}{A.\arabic{figure}}

\setcounter{figure}{0}
\setcounter{table}{0}

{\centerline {\LARGE Supplementary Material}}

\section{Overview of Popular Regularisation Methods}
\label{sec:orgb873690}
First we give a brief overview of the regularisation methods tested. The main goal here is to compare a wide variety of different schemes to see if they have any common behaviour.

\subsection{Weight decay}
\label{sec:org7406e50}
One of the most used methods for regularisation is placing a \(\ell_2\)-penalty on the parameters of the network. Colloquially this is known as \emph{Weight Decay} \citep{krogh91:_simpl_weigh_decay_can_improv_gener}. It is a well understood method from classical regression models, where estimation variance is reduced by regularising large weights or regression coefficients.
%are usually %a symptom of over-fitting certain features in the data.
Despite its easy interpretation for these classical models, its application to neural networks is not completely understood.

% The modified loss with an L2-penalty on the weights is as follows:
% \begin{equation}
% \widetilde{\mathcal{L}} = \mathcal{L} + \alpha ||\boldsymbol{\theta}||_2^2\ .
% \end{equation}

\subsection{Dropout}
\label{sec:orge71dbf0}
\emph{Dropout} \citep{srivastava14:_dropout} is another commonly used regularisation method that works by overlaying noise during the learning procedure. The output from each neuron in the network is discarded at random during training according to a set distribution. Intuitively, by adding this noise, the network is forced to not rely on any given path through the network which could be used to memorise data.

% For a standard feed-forward neural network the output for a neuron in a given hidden layer \(l\) is:
% \begin{align}
% z_i^{l+1} &= \mathbf{w}_i^{l+1} \mathbf{y}^{l} + b_i^{l+1} \\
% y_i^{l+1} &= \sigma(z_i^{l+1})
% \end{align}

% Modifying with dropout yields:
% \begin{align}
% r_i^l &\sim \text{Bernoulli}(p) \\
% \tilde{y}_i^l &= r_i^l y_i^l \\
% z_i^{l+1} &= \mathbf{w}_i^{l+1} \tilde{\mathbf{y}}^{l} + b_i^{l+1} \\
% y_i^{l+1} &= \sigma(z_i^{l+1})
% \end{align}

\subsection{Learning rate algorithms and optimisers}
\label{sec:org5d7f07f}
There exist many optimisers other than standard SGD that also offer regularisation benefits. Optimisers such as Adam \citep{kingma15:_adam} set an adaptive learning rate by using estimations of first and second order terms. On the other simple but effective learning rate algorithms such as \emph{Cyclical Learning Rate} \citep{li20:_cyclic_learn_rate_method_deep_learn_train} simply adjust the learning rate in a more nuanced way that leads to a better learning procedure without using any past training information.
\subsection{Loss gradient norm penalties}
\label{sec:org996cac5}
Loss gradient norm penalties add an additional penalty term to the loss: the norm of the gradient of the loss itself. This will lead the gradients of the gradients being calculated in the final back-propagation step, hence why it is commonly referred to as double back-propagation. First, a back-propagation pass must be done to obtain the gradients, then a second pass to evaluate the weight updates with the gradient norm included. Two versions exist: one which calculates the gradient of the loss with respect to the input data \citep{drucker91:_doubl,hoffman19:_robus_learn_jacob_regul} and another that instead does so with respect to the parameters \citep{barrett21:_implic_gradien_regul,zhao22:_penal_gradien_norm_effic_improv}:

\begin{align*}
\tilde{\mathcal{L}} &= \mathcal{L} + ||\nabla_{x} \mathcal{L}|| \\
\tilde{\mathcal{L}} &= \mathcal{L} + ||\nabla_{\theta} \mathcal{L}||.
\end{align*}

We note one important detail in the relation of the latter, the loss-parameter gradient norm, to the MGS trace penalty:
\begin{equation*}
  ||\nabla_{\theta} \mathcal{L}|| = ||\nabla_{\theta} f \cdot \nabla_f \mathcal{L}|| \leq ||\nabla_{\theta} f||\ ||\nabla_f \mathcal{L}||.
\end{equation*}

Therefore, \(||\nabla_{\theta}L||\) is proportional to the MGS trace penalty as \(||\nabla_{\theta} f||_2^2 = \sqrt{\operatorname{tr} K_{\theta}}\). However depending on how the second term \(||\nabla_f \mathcal{L}||\) behaves the two might not be coupled at all. This might explain why in some instances in the experiments, the loss-parameter gradient penalty performed significantly worse. However, it is still the closest to directly optimising MGS and when the loss-model gradient is well behaved, can perform just as well as MGS. Still, the issue arises as to how this penalty performs when any minima is approached, as the second term will diminish, causing the whole penalty to lose influence over training. This might then lead to the model starting to overfit after landing in a minima, which we have observed in our experiments.

\subsection{Functional regularisation}
\label{sec:org3484548}
Recently it has also been shown that \emph{Kernel Ridge Regression} can be applied to neural networks. This places a penalty on function complexity by trying to minimise the norm of the function given by its \emph{Reproducing Kernel Hilbert Space} (RKHS) (e.g. \citep{bietti19:_kernel_persp_regul_deep_neural_networ}). For neural networks, the NTK can be used to define such a RKHS, however the actual function realised by the network is not necessarily close to its RKHS counterpart. Additionally, in our experience, these penalties are the most costly to calculate and were not efficient enough even to run on small networks.
\section{Connection to Neural Tangent Theory}
\label{sec:orge964373}
It should most certainly be noted that the MGS kernel is indeed the same as the \emph{empirical Neural Tangent Kernel} \(\hat{\Theta}\) which is derived from the \emph{Neural Tangent Kernel} (NTK) itself. From the NTK literature, the NTK describes the evolution of the network function \(f_\theta\) in function space in a kernel gradient descent setting. Therefore we can also make use of findings that have been made from the NTK perspective.

A observation made by about the empirical NTK is the notion of "feature learning speed" \citep{jacot18:_neural_tangen_kernel,baratin21:_implic_regul_neural_featur_align,ronen19:_conver_rate_neural_networ_learn}. Neural networks seem to learn functions/features of increasing complexity starting with low frequency content first and then sequentially higher frequency information as training progresses. The spectral decomposition of the empirical NTK describes the primary directions of learning with large eigenvalues corresponding to learning directions with low complexity which is also where convergence is the fastest. While most of the ideas from the NTK side are formulated by viewing the network as learning in function space, it can be useful to see what practical implications this has on the model gradients and therefore MGS.

%This idea makes sense from the MGS perspective as the largest eigendirections from the spectral decomposition of the MGS kernel, will correspond to the primary gradient directions amongst the input data, which naturally should correspond to directions of low complexity.

Due to the NTK growing in popularity there exist a number of performant libraries for calculating the empirical NTK. We make use of \emph{Neural Tangents} \citep{novak19:_neural_tangen} and \emph{Fast Finite Width NTK} \citep{novak22:_fast_finit_width_neural_tangen_kernel}  libraries which are built on \emph{JAX} [\cite{bradbury18:_jax}].
\section{Using MGS metrics for adversarial sample detection}
\label{sec:org035b763}
The authors \citet{martin20:_inspec}  %inadvertently have already
used \(\operatorname{tr}K_\theta\) for detection of adversarial samples when running a neural network. They do this by calculating the trace of what they call the \emph{approximate Fisher Information Matrix} (FIM) which is the actually same as \(K_{\theta}\) in our case. They observed that if adversarial samples were present in the input, then the trace of the approximate FIM would be large. Their reasoning behind this behaviour, however, was deduced from the properties of the complete FIM. From the gradient similarity perspective it also makes sense. Adversarial samples are intended to look very similar to trained samples, but corrupt the model by causing it to produce different outputs. Therefore, if for adversarial data the trace is large, then the network considers that data to be very different from existing data. This also makes sense as that is one of the objectives of adversarial attacks in the first place.

\section{Experiments}
All experiments using a FCN were run locally on a laptop with a Nvidia Quadro T2000 graphics card. For convolutional networks, the laptop graphics card was not sufficient so instead a single Nvidia K80 was used on another machine.

\subsection{Experiment details}
\paragraph{Classification: MNIST}
\label{sec:experiment_mnist}
A corrupted version (motion blur) of MNIST was used from the MNIST-C dataset \citep{mu19:_mnist_c}. Each method was tuned using a simple grid search over a set of parameters. The scenario in which the tuning took place was using 6000 randomly stratified sampled training points and 50\% of the target labels were randomly flipped. Each method was trained for 100 epochs and also run 5 times over, with differently randomly sample data each time. The final test loss was used to determine the optimal parameter. The architectures tested included a standard fully connected architecture with 6 hidden layers, 300 neurons wide, and ReLU activations as well as a LeNet-5 style convolutional network. Cross-entropy with softmax was used as the loss function. In the case of Dropout, a Dropout layer was added between all fully connected layers in both architectures.

After tuning, each method retained the same parameters and was then run over a large testbench, for 400 epochs for both networks, where the training size and label noise were varied between 250 and 9000 training samples and 0\% to 100\% respectively. The test accuracy was calculated on the pre-determined test sample found in MNIST and not the data that was left over from the training sample. Each scenario was run 10 times, again with new network initialisations and sampled training data. A slow exponential decay was used for the learning rate starting at 0.1 and a batch size of 32 was used.

Each of the metrics, including test accuracy, were sampled at 100 evenly spaced points during training, so as to not run too slowly. The final test accuracy was calculated as an average of the previous 5 metric samples and then averaged between the training runs. The max value was calculated as the max of the averaged runs.

\paragraph{Regression: Facial Keypoints}
A similarly corrupted version (motion blur) of the Facial Keypoints dataset \footnote[1]{https://www.kaggle.com/competitions/facial-keypoints-detection/} was used. The same architecture types were used as in the MNIST experiment. Standard MSE loss is used. Unlike the MNIST experiment, a grid of scenarios was not run for this problem. Instead the training size was held fixed at 30\% of the total data, while the amount of noise was changed. Gaussian noise was added to the target coordinates with a scale indicated by the ``noise'' column in the tables. Each method was tuned with a noise scale of 15 for 50 epochs.

Each scenario was then run for 400 epochs in the case of the FCN architecture and 50 epochs for the LeNet-5 architecture. This was due to the larger size of the input and output, causing the convolutional network to run much slower. A batch size of 128 was used for both architectures in an attempt to lessen the training wall-clock time. This could explain why MGS in some cases achieved slightly better performance for the FCN architecture over the LeNet-5 architecture as it was allowed to train for longer.

The metrics were tracked in the same way as the MNIST experiment, however test MSE was used instead of test accuracy.

\paragraph{Training parameter robustness}
To test robustness with regards to different training parameters, a benchmark was run where each method is first tuned using the same scenario. Then, the tuned scenario is used as a starting point from which each of the 5 training parameters were changed independently. Both larger and smaller parameter values were selected to test as broad of a spectrum as possible. For each parameter changed, multiple runs were performed, again by resampling the data and using a different initialisation. The final performance results were aggregated, with the quantiles being drawn, based on all the results. The test was run using the same classification problem based on the corrupted MNIST dataset described previously. Specifically, each method was tuned using 3000 training samples, 50\% label noise, and for 100 epochs with all the other settings such as learning rate the same as per section \ref{sec:experiment_mnist}.

\subsection{Results for MNIST and Facial Keypoints for a FCN network}

\FloatBarrier

\begin{table}
  \renewcommand\thetable{A.1}
  \caption{\textbf{Test accuracy for corrupted MNIST dataset using a FCN architecture.} Noise column represents percentage of training labels that have been randomly flipped.  MGS is able to handle large amounts of noise. If the max accuracy is compared to the final accuracy, MGS is also the most consistent and is not susceptible to overfitting. The FCC results are worse than those of LeNet-5. }
  \label{tbl:perf-mnist-fcn}
  \centering
  \begin{tabular}{cccccc}
    \toprule
    % \multicolumn{5}{c}{Gradient penalty (parameter)}                   \\
    % \cmidrule(r){1-2}
    % Name     & Description     & Size ($\mu$m) \\
    Noise & Unregularised & Dropout & Weight & Loss grad. & MGS \\
    \midrule
    0\% & \parbox{5em}{\centering 82.1 \textpm 1.9 \\ (82.3)} & \parbox{5em}{\centering 79.5 \textpm 1.2 \\ (80.0)} & \parbox{5em}{\centering 72.8 \textpm 9.7 \\ (79.6)} & \parbox{5em}{\centering 74.9 \textpm 22.0 \\ (79.8)} & \parbox{5em}{\centering \textbf{84.6} \textpm 1.2 \\ (84.8)} \\
    \addlinespace
    30\% & \parbox{5em}{\centering 59.5 \textpm 3.7 \\ (75.3)} & \parbox{5em}{\centering 73.7 \textpm 2.5 \\ (75.5)} & \parbox{5em}{\centering 58.7 \textpm 4.0 \\ (74.6)} & \parbox{5em}{\centering 70.3 \textpm 4.3 \\ (77.4)} & \parbox{5em}{\centering \textbf{77.7} \textpm 1.8 \\ (79.4)} \\
    \addlinespace
    60\% & \parbox{5em}{\centering 36.9 \textpm 2.8 \\ (65.6)} & \parbox{5em}{\centering 55.6 \textpm 4.8 \\ (57.1)} & \parbox{5em}{\centering 10.0 \textpm 0.5 \\ (63.1)} & \parbox{5em}{\centering 41.2 \textpm 6.4 \\ (43.4)} & \parbox{5em}{\centering \textbf{67.8} \textpm 3.4 \\ (70.0)}\\
    \addlinespace
    80\% & \parbox{5em}{\centering 24.8 \textpm 2.7 \\ (53.0)} & \parbox{5em}{\centering 26.5 \textpm 3.7 \\ (29.6)} & \parbox{5em}{\centering 10.1 \textpm 0.6 \\ (41.3)} & \parbox{5em}{\centering 12.1 \textpm 2.5 \\ (14.8)} & \parbox{5em}{\centering \textbf{51.4} \textpm 3.8 \\ (55.0)}\\
  \bottomrule
  \end{tabular}
\end{table}

\FloatBarrier

\begin{table}
\renewcommand\thetable{A.2}
  \caption{\textbf{Test loss for the corrupted Facial Keypoints dataset using a FCN architecture.} MGS is the only regulariser that is able to attain convergence with good performance.}
  \label{tbl:perf-fk-fcn}
  \centering
  \begin{tabular}{cccccc}
    \toprule
    Noise & Unregularised & Dropout & Weight & Loss grad. & MGS\\
    \midrule
    0 & \parbox{5.2em}{\centering 281.1 \textpm 264.6 \\ (181.5)} & \parbox{5.2em}{\centering 208.2 \textpm 220.9 \\ (74.3)} & \parbox{5.2em}{\centering 276.4 \textpm 262.3 \\ (180.2)} & \parbox{5.2em}{\centering 337.9 \textpm 303.8 \\ (184.7)} & \parbox{5.2em}{\centering \textbf{10.6} \textpm 0.3 \\ (10.5)}\\
    \addlinespace
    10 & \parbox{5.2em}{\centering 305.7 \textpm 344.7 \\ (176.8)} & \parbox{5.2em}{\centering 240.2 \textpm 291.1 \\ (191.7)} & \parbox{5.2em}{\centering 305.2 \textpm 342.2 \\ (175.0)} & \parbox{5.2em}{\centering 335.3 \textpm 332.1 \\ (208.2)} & \parbox{5.2em}{\centering \textbf{11.0} \textpm 0.5 \\ (10.7)}\\
    \addlinespace
    20 & \parbox{5.2em}{\centering 325.2 \textpm 371.7 \\ (158.6)} & \parbox{5.2em}{\centering 287.1 \textpm 321.9 \\ (88.0)} & \parbox{5.2em}{\centering 328.5 \textpm 379.4 \\ (157.7)} & \parbox{5.2em}{\centering 353.1 \textpm 382.7 \\ (174.2)} & \parbox{5.2em}{\centering \textbf{13.6} \textpm 1.3 \\ (12.8)}\\
    \addlinespace
    30 & \parbox{5.2em}{\centering 390.8 \textpm 453.2 \\ (162.6)} & \parbox{5.2em}{\centering 210.4 \textpm 270.9 \\ (91.7)} & \parbox{5.2em}{\centering 393.0 \textpm 454.2 \\ (162.2)} & \parbox{5.2em}{\centering 411.8 \textpm 442.8 \\ (175.6)} & \parbox{5.2em}{\centering \textbf{22.4} \textpm 2.7 \\ (20.9)}\\
    \bottomrule
  \end{tabular}
\end{table}

\FloatBarrier

\section{Two circles problem decision boundaries}

\FloatBarrier

\begin{figure}
  \includegraphics[width=\textwidth]{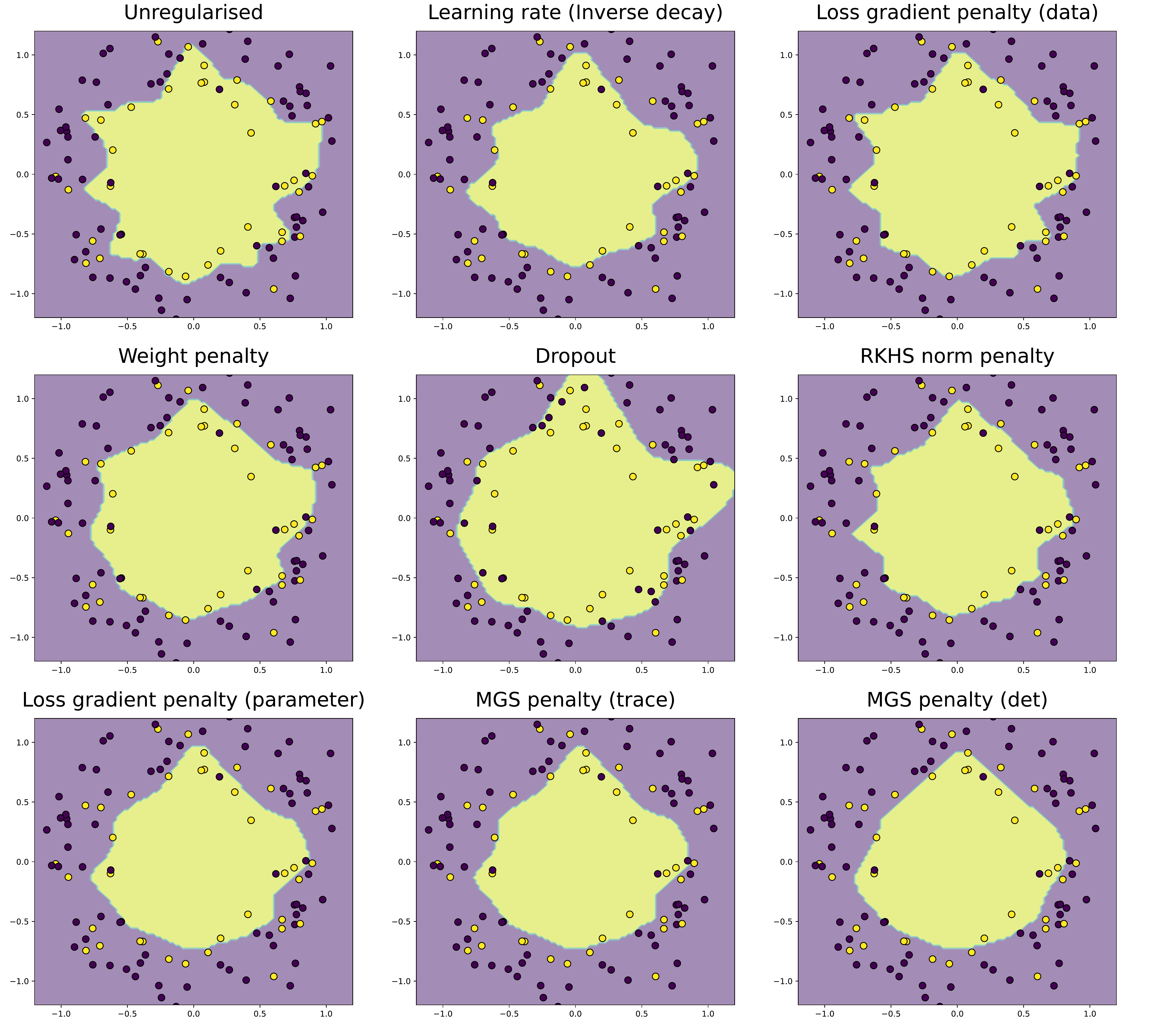}
  \caption{Decision boundaries for all regularisation methods.}
\end{figure}

\FloatBarrier

\clearpage

\section{MNIST testbench results}

\FloatBarrier

\begin{figure}
\renewcommand{\thefigure}{A.2}
  \includegraphics[width=\textwidth]{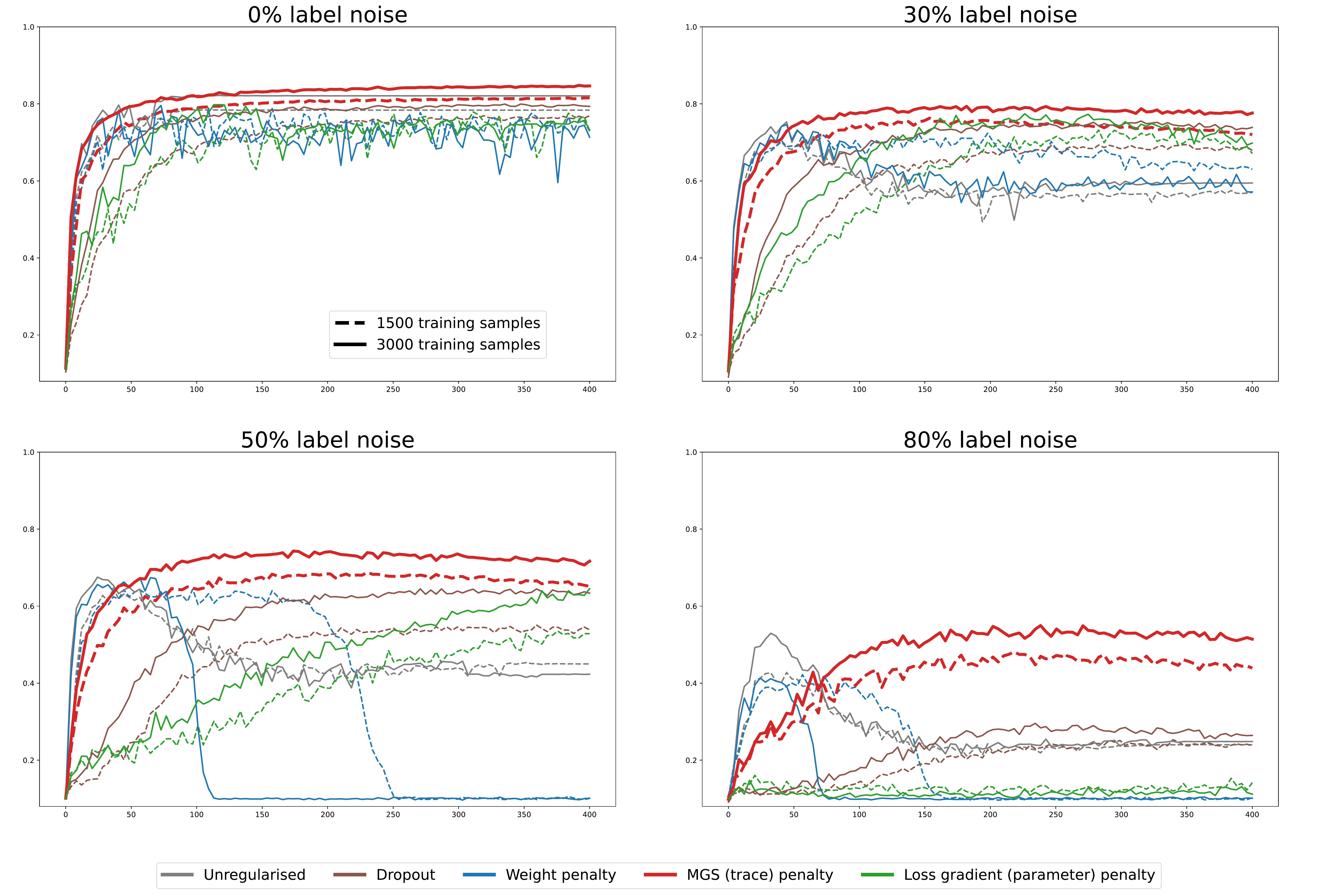}
  \caption{\textbf{Test accuracy during training using a FCN architecture.} Test accuracy plotted during training for two training sizes and different levels of label noise.}
\end{figure}

\FloatBarrier

\begin{figure}
\renewcommand{\thefigure}{A.3}
  \includegraphics[width=\textwidth]{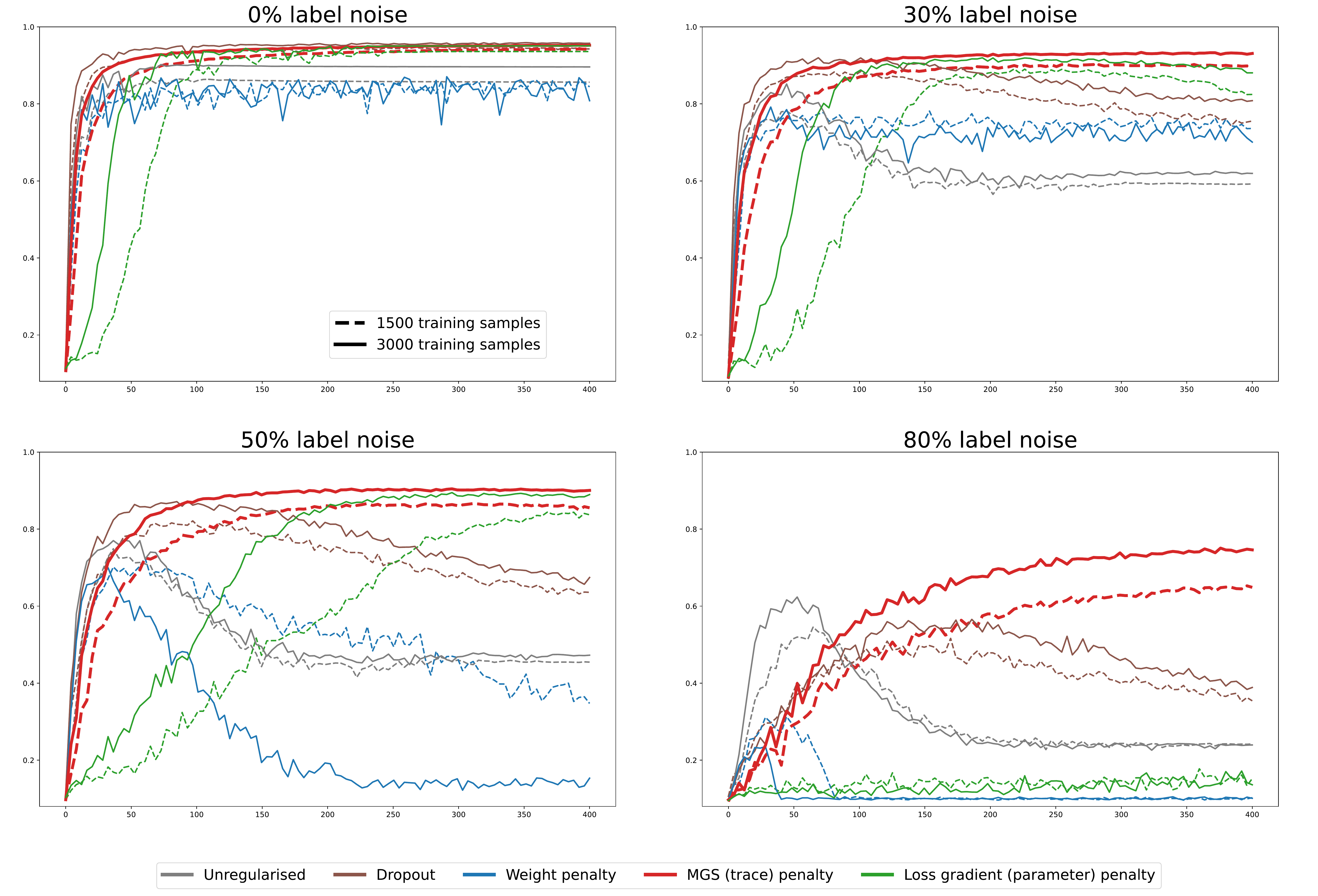}
  \caption{\textbf{Test accuracy during training using a LeNet-5 architecture.} Test accuracy plotted during training for two training sizes and different levels of label noise.}
\end{figure}

\FloatBarrier

\begin{figure}
\renewcommand{\thefigure}{A.4}
  \includegraphics[width=\textwidth]{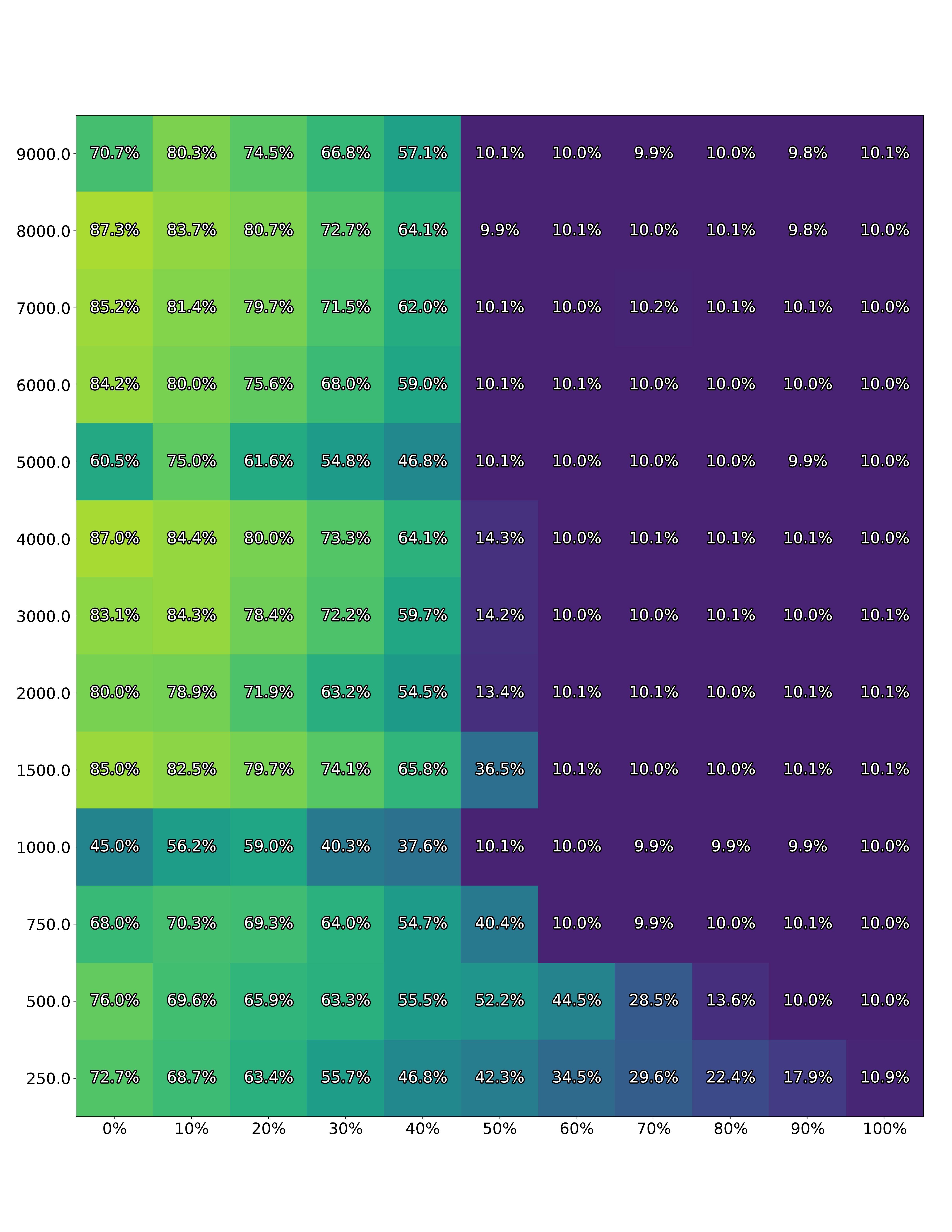}
  \caption{\textbf{Unregularised LeNet-5}. Average final test accuracy for MNIST. Label noise on the x-axis and training size on the y-axis.}
\end{figure}

\FloatBarrier

\begin{figure}[p]
\renewcommand{\thefigure}{A.5}
  \includegraphics[width=\textwidth]{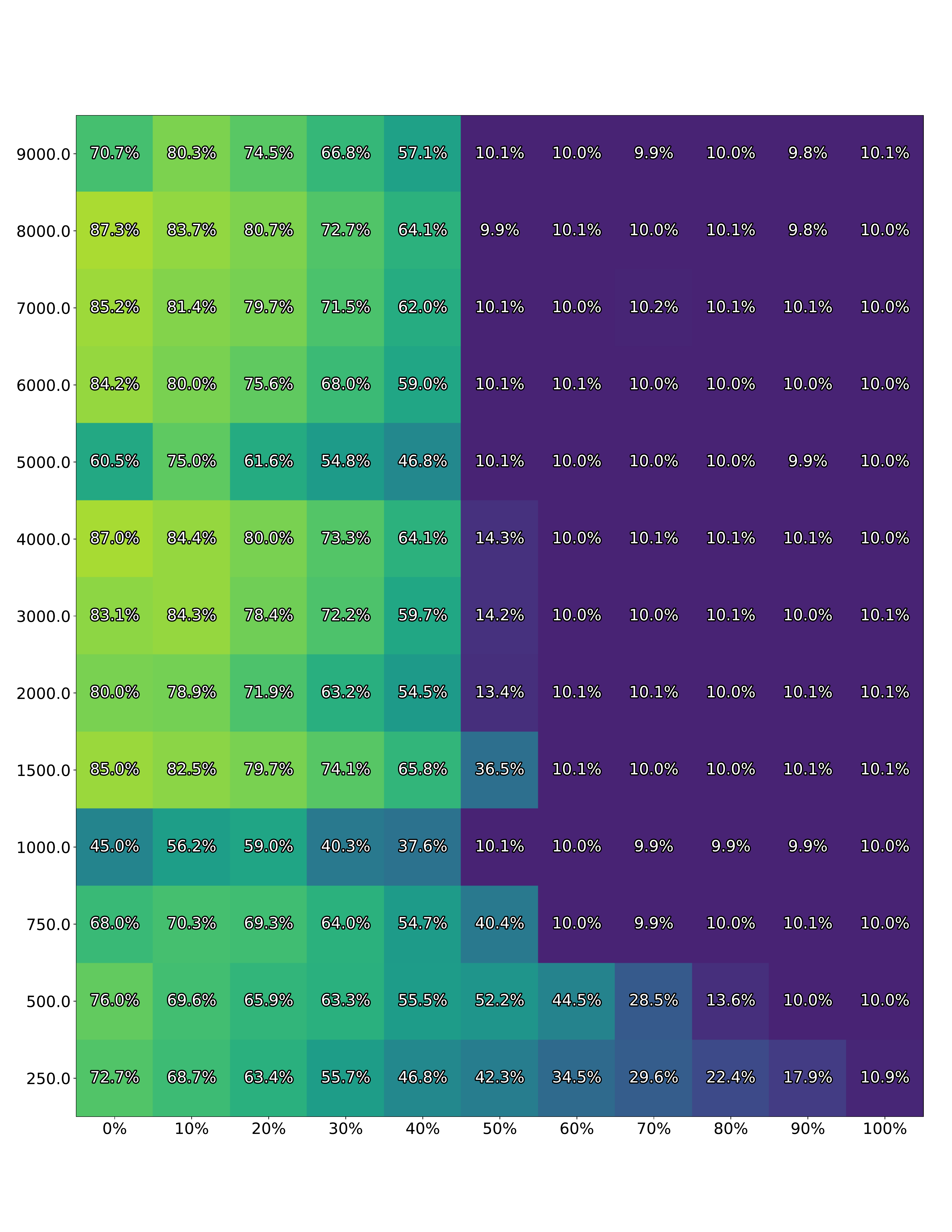}
  \caption{\textbf{LeNet-5 regularised with weight penalty}. Average final test accuracy with label noise on the x-axis and training size on the y-axis.}
\end{figure}

\FloatBarrier

\begin{figure}[p]
\renewcommand{\thefigure}{A.6}
  \includegraphics[width=\textwidth]{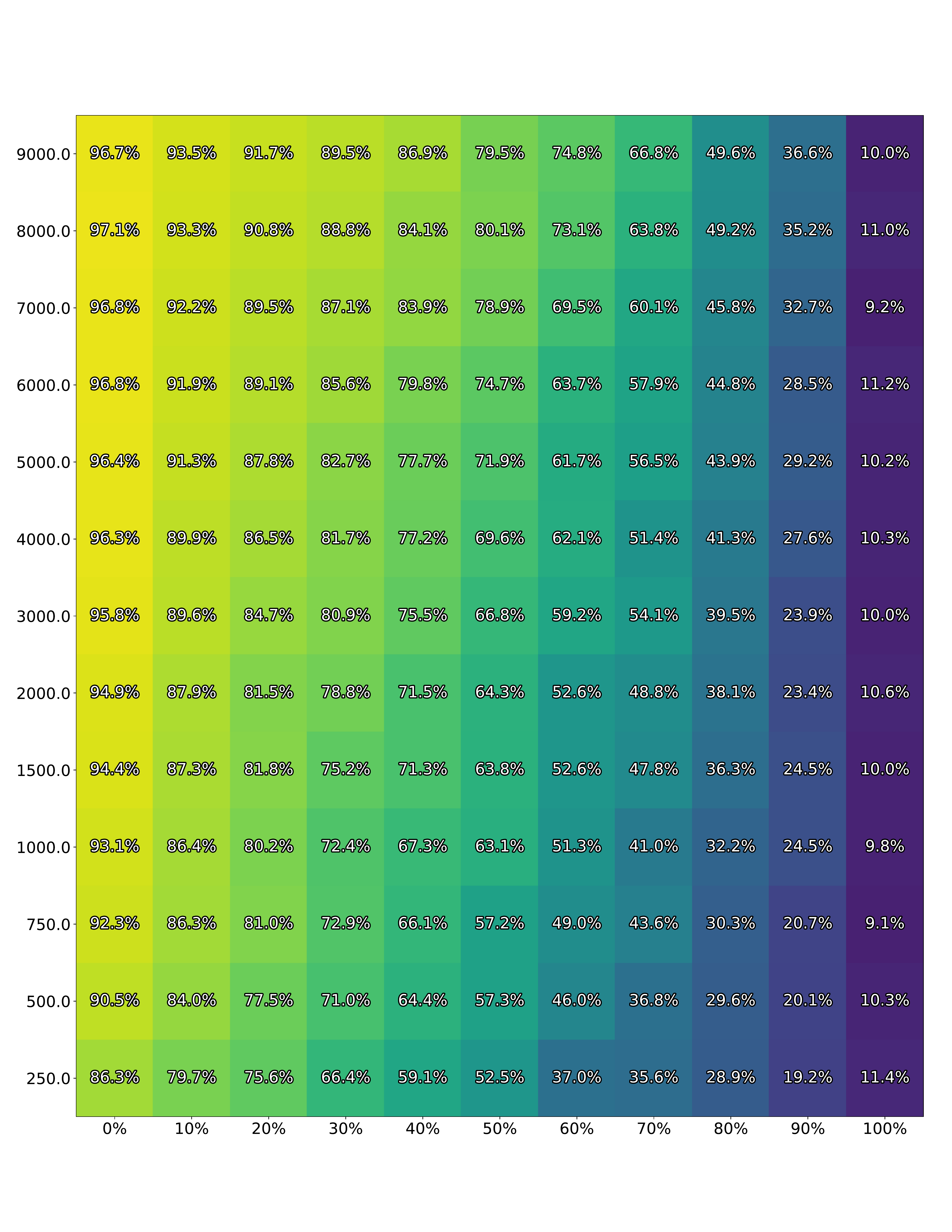}
  \caption{\textbf{LeNet-5 regularised with dropout}. Average final test accuracy with label noise on the x-axis and training size on the y-axis.}
\end{figure}

\FloatBarrier

\begin{figure}[p]
\renewcommand{\thefigure}{A.7}
  \includegraphics[width=\textwidth]{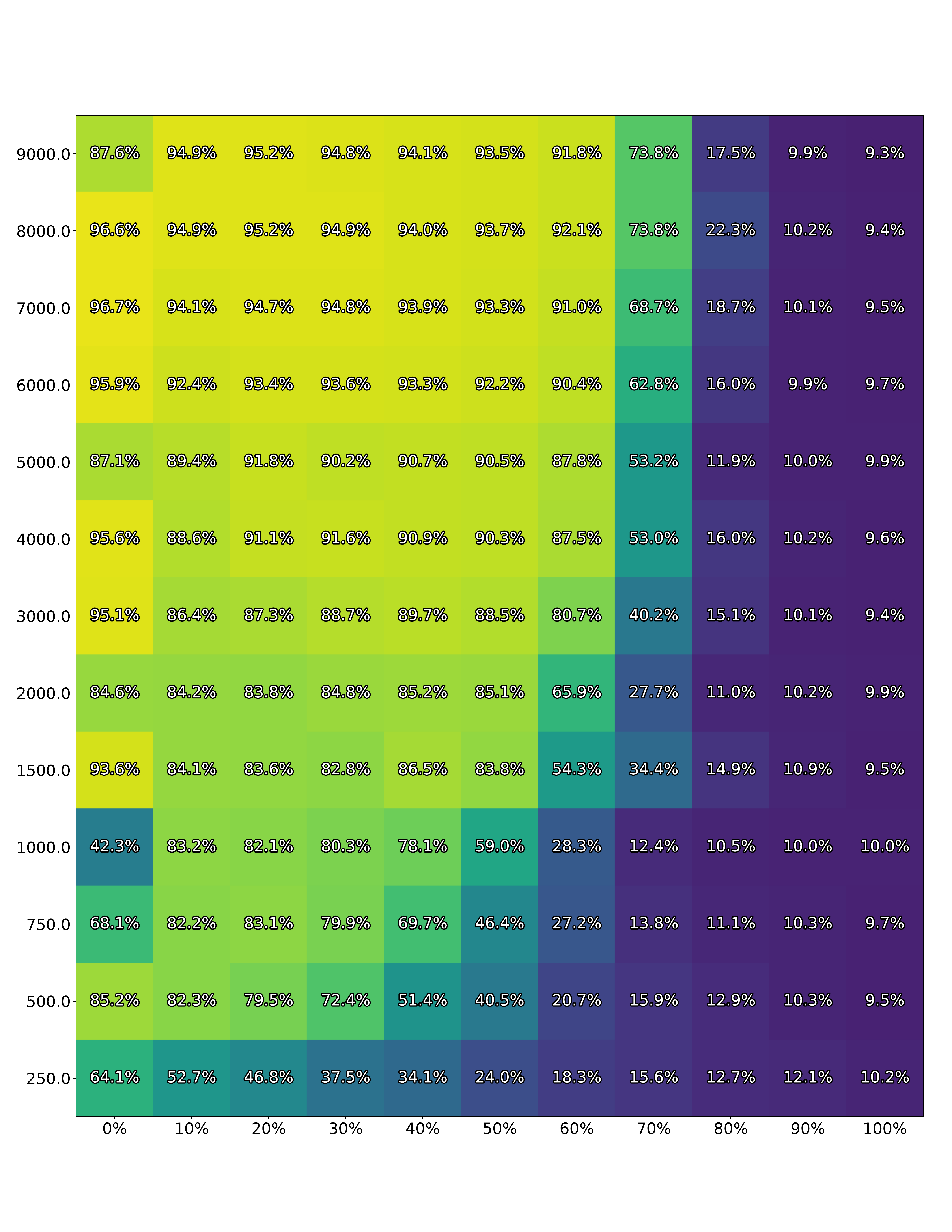}
  \caption{\textbf{LeNet-5 regularised with loss-gradient penalty (parameter)}. Average final test accuracy with label noise on the x-axis and training size on the y-axis.}
\end{figure}

\FloatBarrier

\begin{figure}[p]
\renewcommand{\thefigure}{A.8}
  \includegraphics[width=\textwidth]{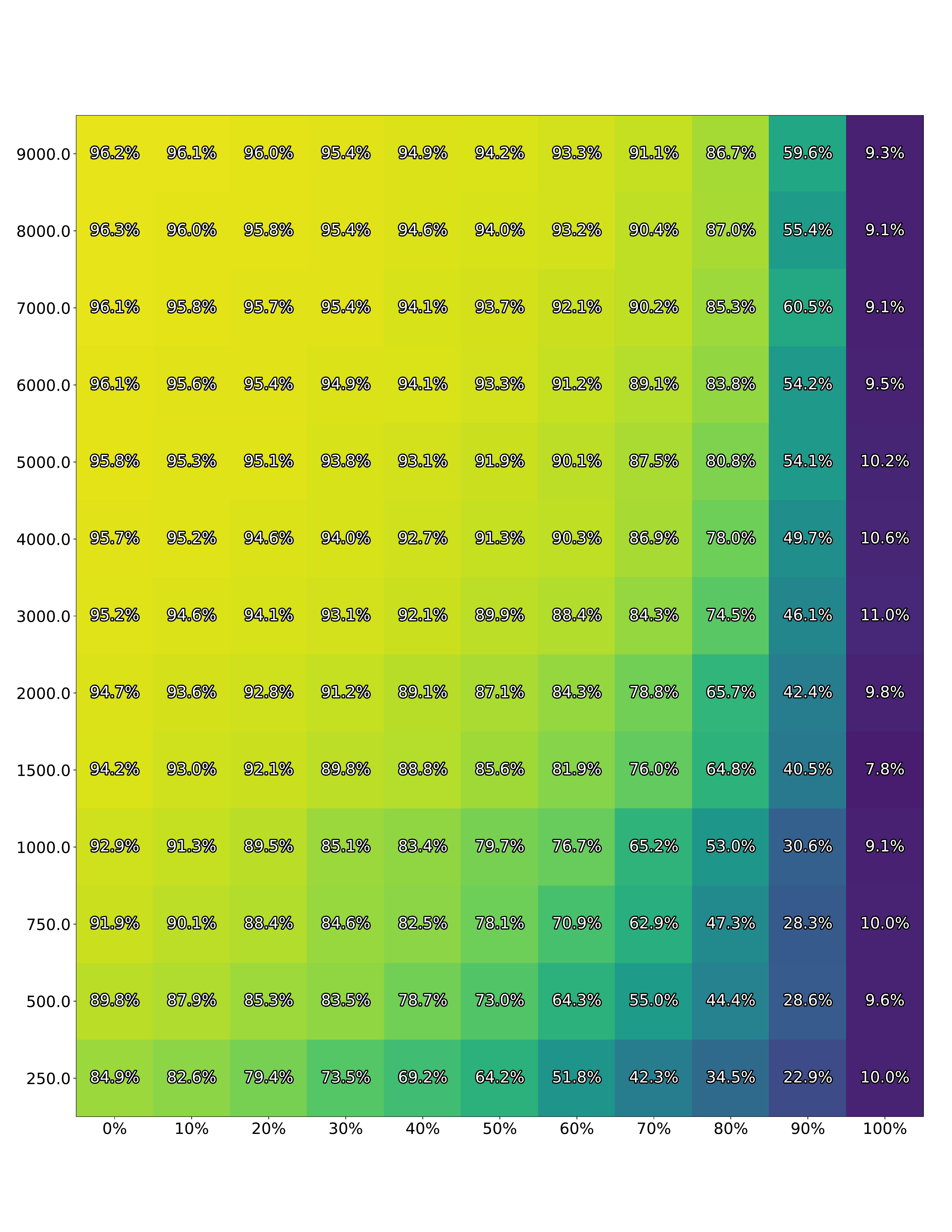}
  \caption{\textbf{LeNet-5 regularised with MGS penalty (trace)}. Average final test accuracy with label noise on the x-axis and training size on the y-axis.}
\end{figure}

\FloatBarrier

\begin{figure}[p]
\renewcommand{\thefigure}{A.9}
  \includegraphics[width=\textwidth]{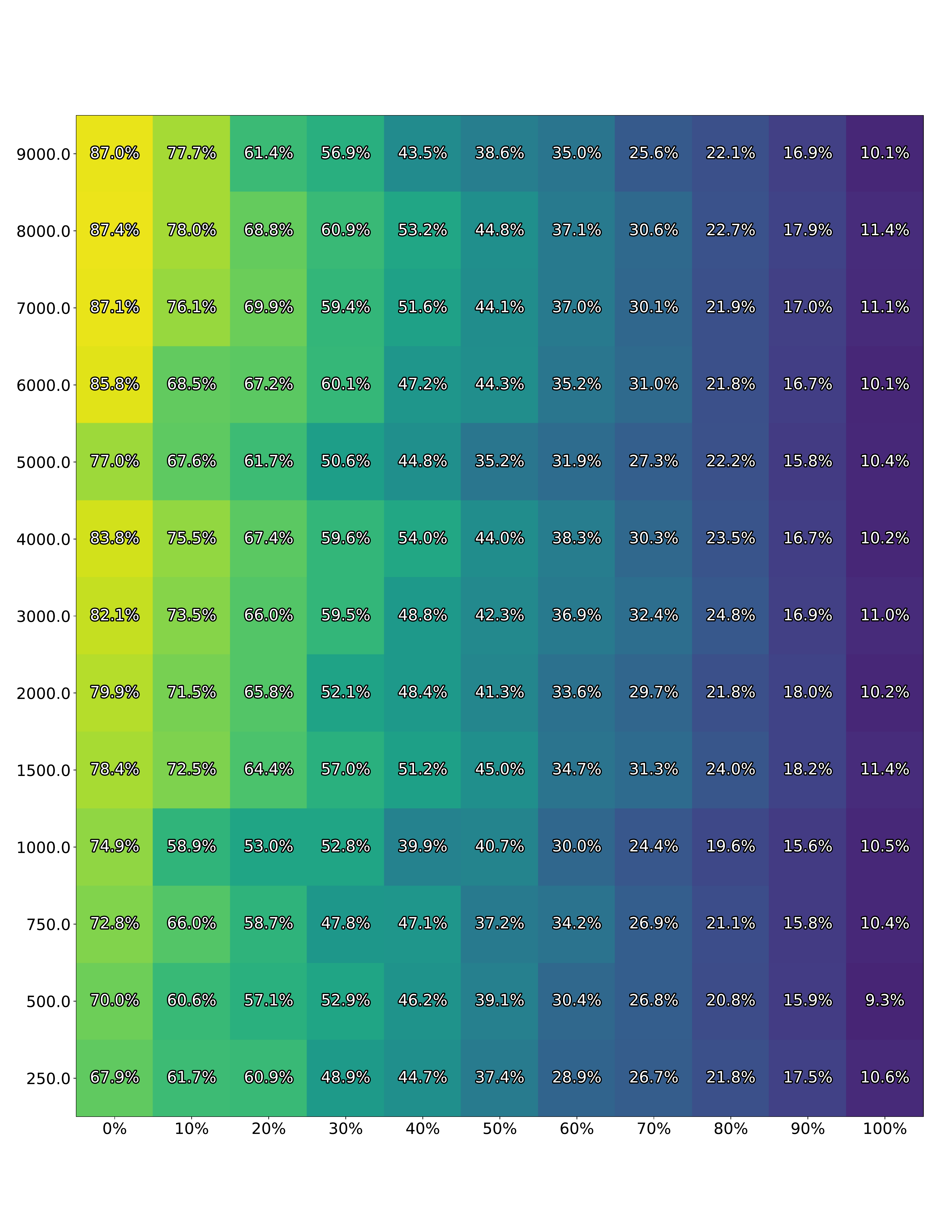}
  \caption{\textbf{Unregularised FCN}. Average final test accuracy for MNIST. Label noise on the x-axis and training size on the y-axis.}
\end{figure}

\FloatBarrier

\begin{figure}[p]
\renewcommand{\thefigure}{A.10}
  \includegraphics[width=\textwidth]{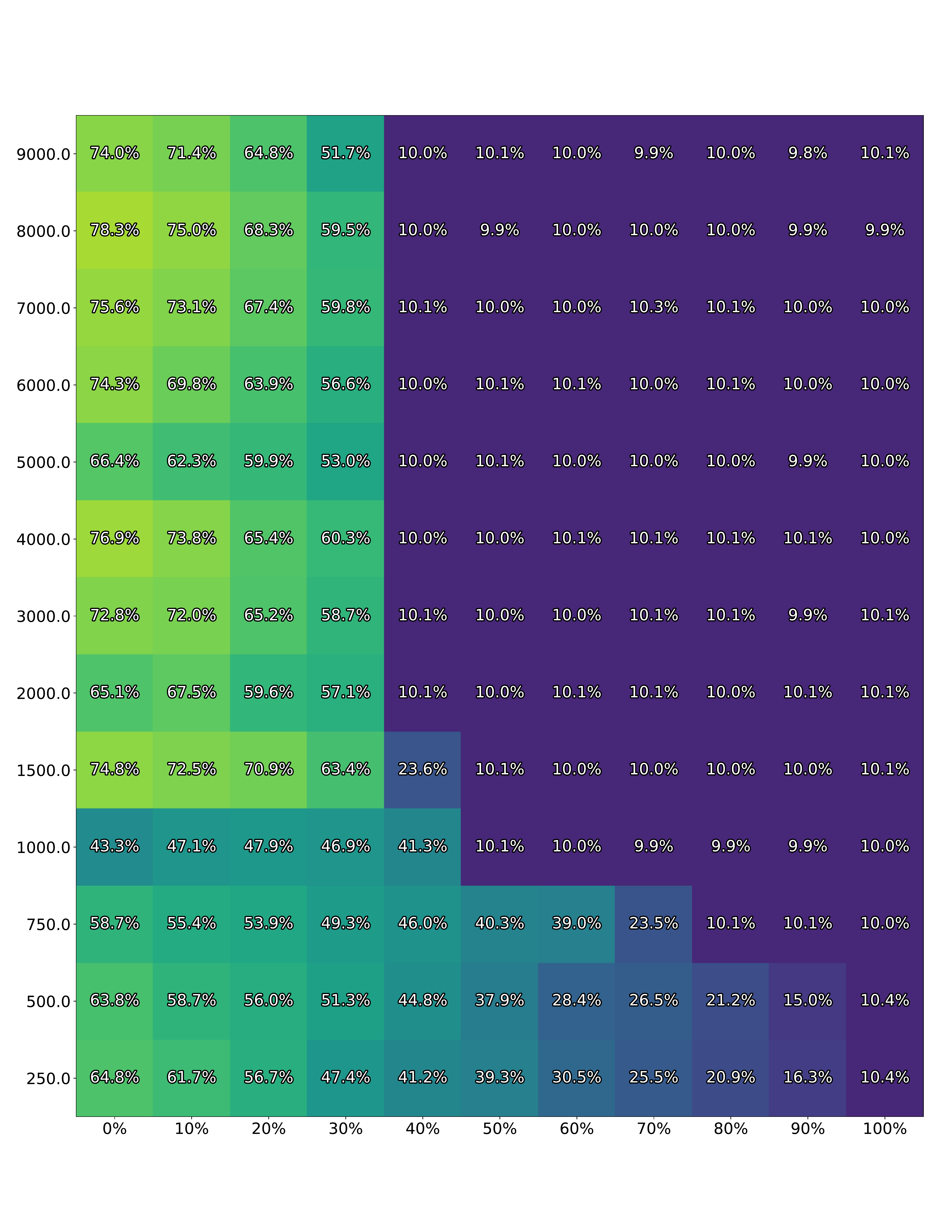}
  \caption{\textbf{FCN regularised with weight penalty}. Average final test accuracy with label noise on the x-axis and training size on the y-axis.}
\end{figure}

\FloatBarrier

\begin{figure}[p]
\renewcommand{\thefigure}{A.11}
  \includegraphics[width=\textwidth]{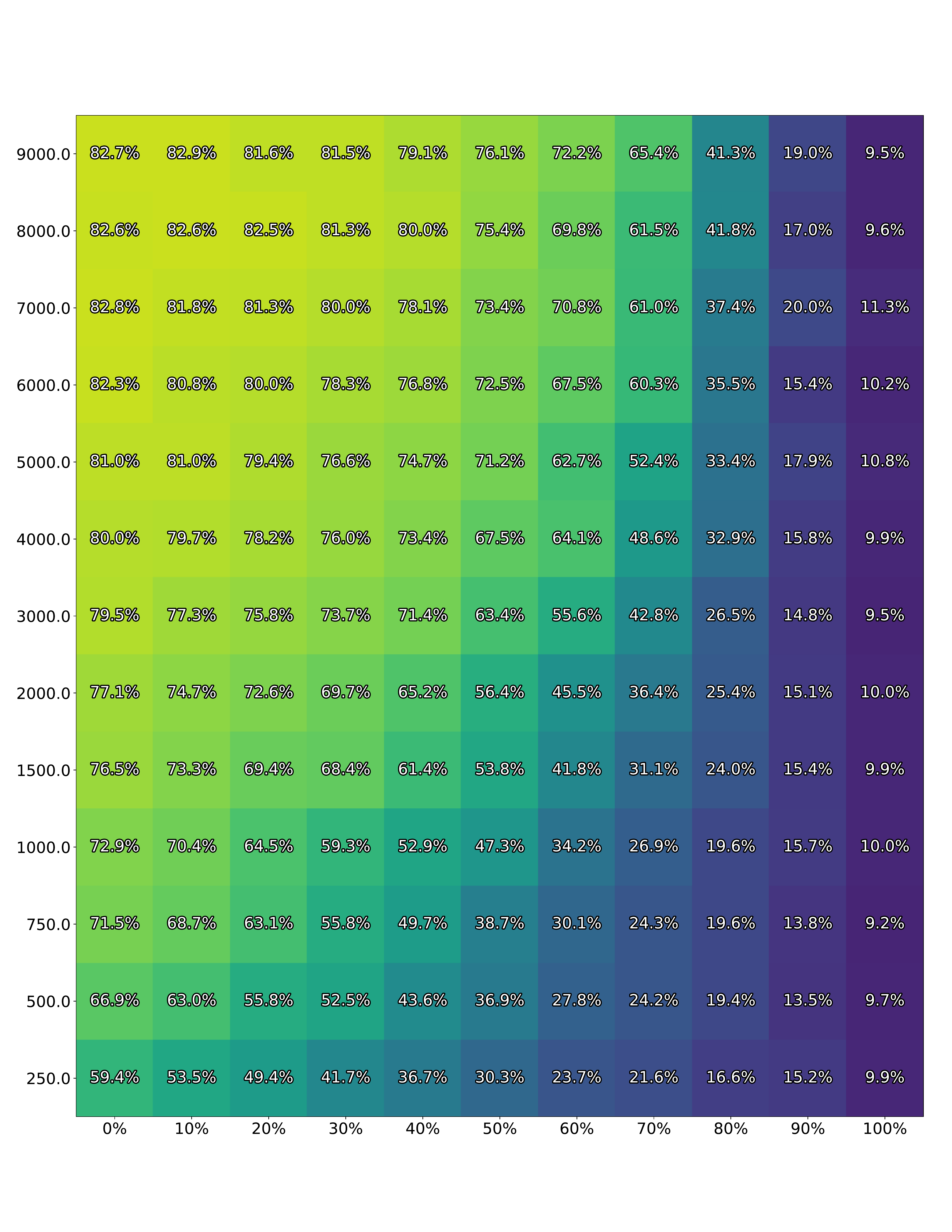}
  \caption{\textbf{FCN regularised with dropout}. Average final test accuracy with label noise on the x-axis and training size on the y-axis.}
\end{figure}

\FloatBarrier

\begin{figure}[p]
\renewcommand{\thefigure}{A.12}
  \includegraphics[width=\textwidth]{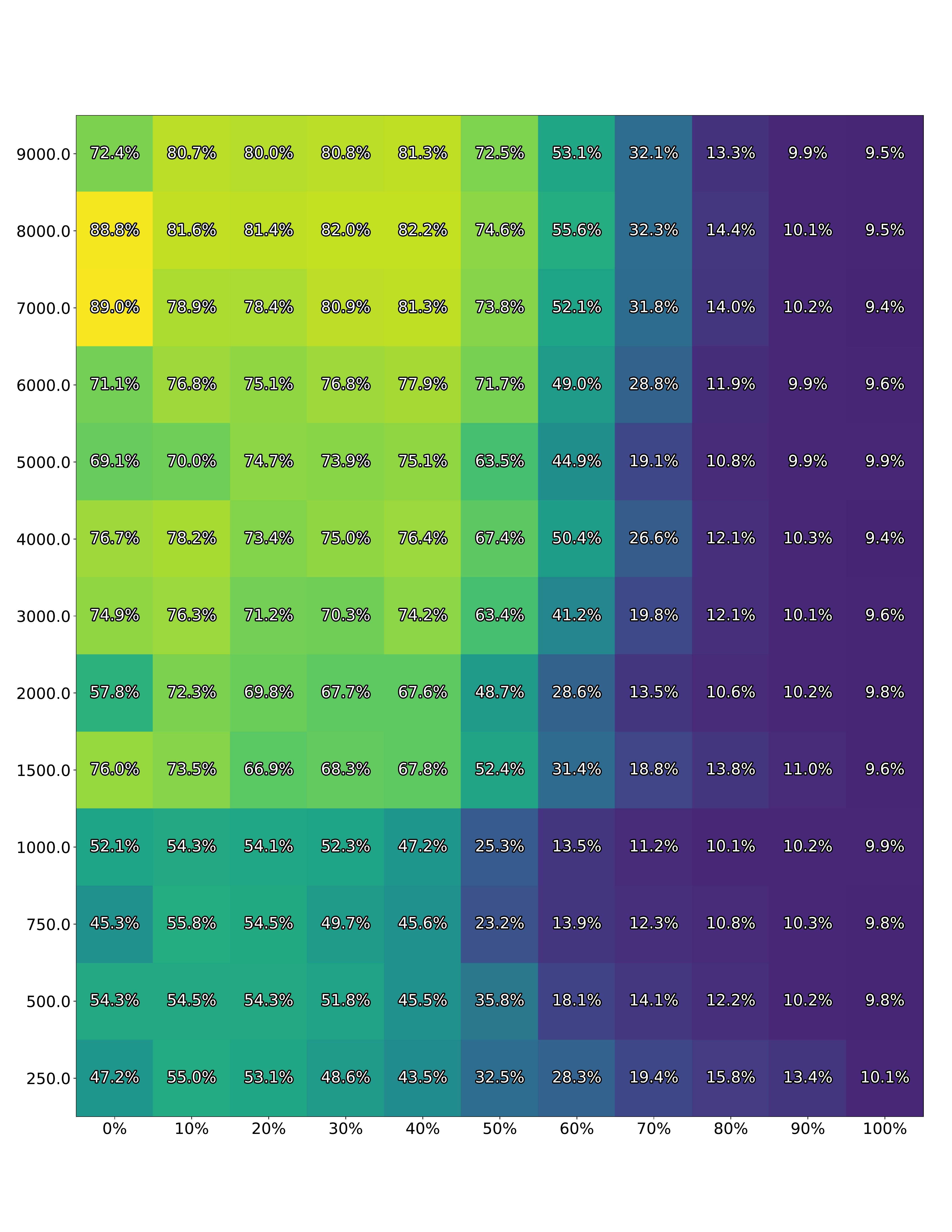}
  \caption{\textbf{FCN regularised with loss-gradient penalty (parameter)}. Average final test accuracy with label noise on the x-axis and training size on the y-axis.}
\end{figure}

\FloatBarrier

\begin{figure}[p]
\renewcommand{\thefigure}{A.13}
  \includegraphics[width=\textwidth]{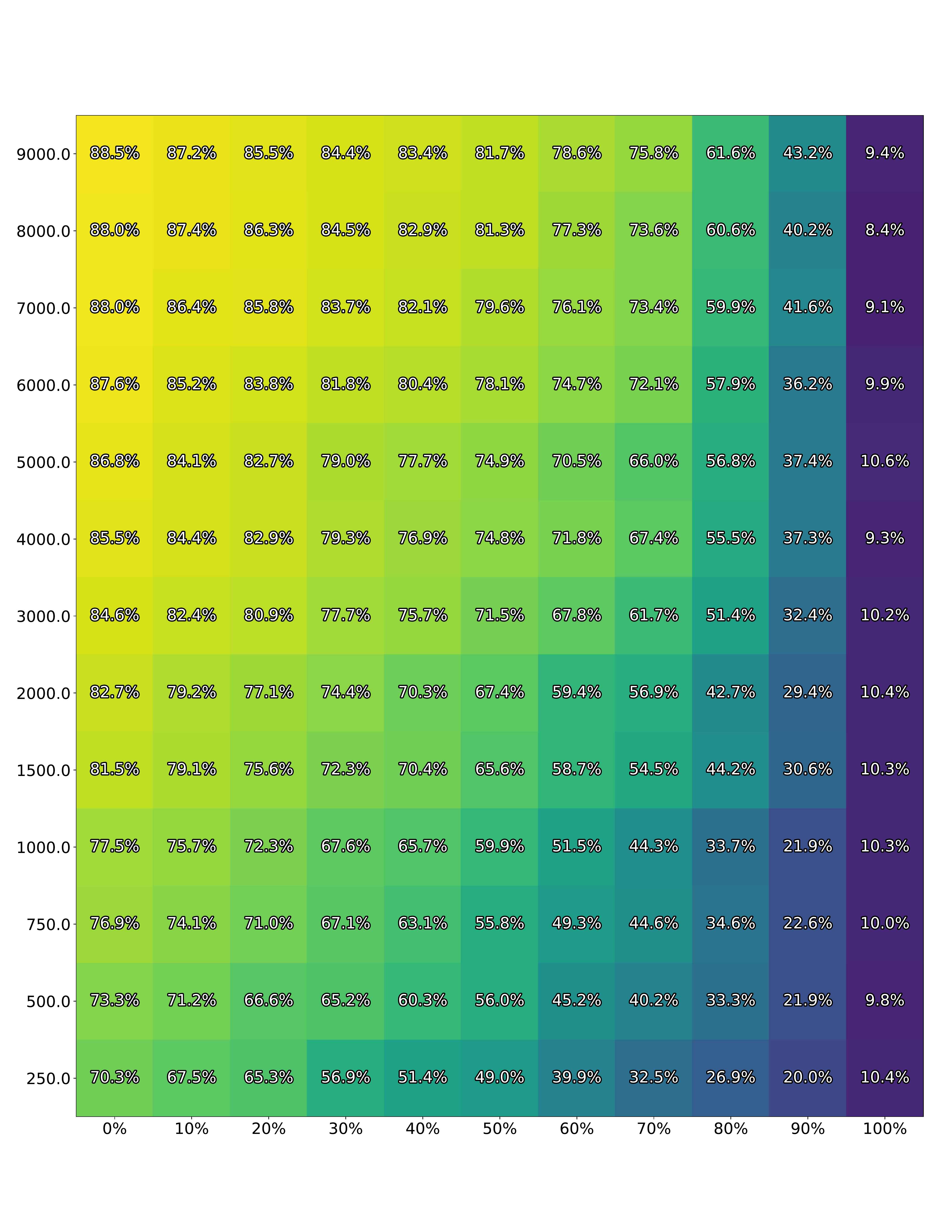}
  \caption{\textbf{FCN regularised with MGS penalty (trace)}. Average final test accuracy with label noise on the x-axis and training size on the y-axis.}
\end{figure}

\FloatBarrier

\section{Calculation of \(K_{\theta}\)}
\subsection{Vector output models}
In section \ref{sec:mgs} \(K_{\theta}\) is defined for a scalar output model and therefore is a matrix of size \(n \times n\). In the case of a vector output model \(K_{\theta}\) becomes a tensor of size \(n \times n \times q \times q\) where \(q\) is the number of outputs. The reasoning about MGS still holds. However, the calculations become more involved as one not only has to consider the model update for a single output (given by the diagonal matrices \(K^{i,j}_{\theta}, i = j \)) but also how an update for one output affects other outputs (given by the off diagonal matrices \(K^{i,j}_{\theta}, i \neq j \)). Therefore, it is easier to instead combine all the outputs by concatenating their model gradients and then calculate a single kernel \(K_{\theta}\) of size \(np \times np\), as is common in the NTK literature \citep{lee19:_wide_neural_networ_any_depth}:
\begin{equation*}
  \nabla_\theta f = [\nabla_{\theta} f_1; \cdots; \nabla_{\theta} f_q ].
\end{equation*}
%This is equivalent to taking the tensor contraction of \(K_{\theta}\), first by summing the diagonal matrices \(K^i_{\theta}= \sum_i^q K^{i,i}_{\theta} \), and then calculating the trace of \(K^i_{\theta}\)

\subsection{Handling of mini-batches and large data sets}
When using mini-batches, \(K_{\theta}\) is no longer square and instead will be of size \(n \times m\), where \(m\) is the number of samples in the current batch.

Both this matrix and the \(n \times n\) matrix from standard gradient descent can be infeasible to calculate for large data sets. Instead, an approximation is calculated using only the data from the current batch which will be of size \(m \times m\). Throughout the entire article, \(K_{\theta}\) is always calculated for the current batch and not the full dataset.

\section{Relation between spectrum of MGS kernel and gradient sample covariance matrix}

\begin{prop}
  \label{prop:gram}
  The matrices \(X^TX\) (Gram matrix) and \(XX^T\) (scatter matrix) share the same non-zero eigenvalues.

  Proof: Suppose that \(\lambda\) is a non-zero eigenvalue of \(X^TX\) with the associated eigenvector u.

  Then:
  \begin{align*}
    X^TXu &= \lambda u \\
    XX^TXu &= X\lambda u \\
    XX^T(Xu) &= \lambda (Xu) \\
    XX^T \tilde{u} &= \lambda \tilde{u}.
  \end{align*}

  Thus, \(\lambda\) is an eigenvalue of \(XX^T\), with the associated eigenvector \(\tilde{u} = XU. \blacksquare\)
\end{prop}

\begin{thm} Separation theorem \citep{takane91:_princ}.
  \label{thm:separation}

  Let \(M\) be a \(d\)-by-\(n\) matrix. Let two orthogonal projection matrices be \(P_{\text{left}}\) and \(P_{\text{right}}\) of size \(d\)-by-\(d\) and \(n\)-by-\(n\) respectively.

  Then:
  \begin{equation*}
    \sigma_{j+t}(M) \leq \sigma_j(P_{\text{left}} M P_{\text{right}}) \leq \sigma_{j}(M),
  \end{equation*}
  where \(\sigma_j(\cdot)\) denotes the j-th largest singular value of the matrix, while \(t=d - r(P_{\text{left}}) + n - r(P_{\text{right}})\) and \(r(\cdot)\) is the rank of the matrix.
\end{thm}

\begin{thm}
  \label{thm:interlace}
  Let \(C\) and \(\bar{C}\) be the scatter matrix and its centred counterpart, then their eigenvalues are interlaced, such that \citep{honeine14}:

  \begin{equation*}
    \lambda_{j+1} \leq \bar{\lambda}_{j} \leq \lambda_{j}.
  \end{equation*}

  Proof: Let \(M = X\), \(P_{\text{left}}\) being the \(d\)-by-\(d\) identity matrix and \(P_{\text{right}} = (I - \frac{1}{n}11^T)\) which is the \(n\)-by-\(n\) centering matrix.

  With \(r(P_{\text{left}}) = d\) and \(r(P_{\text{right}}) = n-1\), it follows from the Separation theorem \ref{thm:separation} that:
  \begin{equation*}
    \sigma_{j+1}(X) \leq \sigma_{j}(X - (I - \frac{1}{n}11^T)) \leq \sigma_{j} (X)
  \end{equation*}

  Furthermore it is well known that the eigenvalues of \(C\) are equal to the square roots of the singular values of \(X. \blacksquare\)

\end{thm}

By using proposition \ref{prop:gram} and theorem \ref{thm:interlace} we see that the eigenvalues of the MGS kernel, which is the Gram matrix for the model-gradients, and the centred scatter matrix (sample-covariance matrix) will be interlaced.

\end{document}